%% file: arxiv.tex
\documentclass[letterpaper,10pt,conference]{ieeeconf}
\IEEEoverridecommandlockouts
\input{def}

\usepackage[utf8]{inputenc}
\usepackage[T1]{fontenc}
\usepackage{microtype}
\usepackage{cite}
\usepackage{etoolbox}
\providecommand{\refname}{References}
\patchcmd{\thebibliography}{\section*{References}}{\section*{\refname}}{}{}
\patchcmd{\thebibliography}{\addcontentsline{toc}{section}{References}}{\addcontentsline{toc}{section}{\refname}}{}{}

\usepackage{amsmath,amssymb,amsfonts,amsthm}
\usepackage{graphicx}
\usepackage{textcomp}
\usepackage{booktabs}
\usepackage{multirow}
\let\labelindent\relax
\usepackage{enumitem}
\usepackage[table]{xcolor}
\definecolor{paperlinkcolor}{RGB}{120,127,208}
\definecolor{paperurlcolor}{RGB}{209,166,198}
\usepackage{url}
\usepackage{xspace}
\usepackage{pifont}
\usepackage{algorithm}
\usepackage{algorithmic}
\usepackage{float}
\usepackage{array}
\makeatletter
\let\NAT@parse\@undefined
\makeatother
\usepackage[colorlinks=true,
            linkcolor=paperlinkcolor,
            citecolor=paperlinkcolor,
            urlcolor=paperurlcolor]{hyperref}
\makeatletter
\renewcommand{\IEEEaftertitletext}[1]{\def\@IEEEaftertitletext{#1}}
\newlength{\papertopfloatoffset}
\AtBeginDocument{%
    \settoheight{\papertopfloatoffset}{\normalfont\normalsize H}%
    \setlength{\papertopfloatoffset}{\dimexpr\topskip-\papertopfloatoffset\relax}%
    \patchcmd{\@cflt}{\unvbox\@tempboxa}%
        {\vskip\papertopfloatoffset\unvbox\@tempboxa}{}%
        {\PackageError{paper}{Single-column float alignment patch failed}{}}%
    \patchcmd{\@cflt}{\vskip\textfloatsep}%
        {\vskip\textfloatsep\vskip-\papertopfloatoffset}{}%
        {\PackageError{paper}{Single-column float spacing patch failed}{}}%
    \patchcmd{\@combinedblfloats}{\unvbox\@tempboxa}%
        {\ifnum\@dbltopnum>\m@ne\vskip\papertopfloatoffset\fi\unvbox\@tempboxa}{}%
        {\PackageError{paper}{Double-column float alignment patch failed}{}}%
    \patchcmd{\@combinedblfloats}{\vskip\dbltextfloatsep}%
        {\vskip\dbltextfloatsep\ifnum\@dbltopnum>\m@ne\vskip-\papertopfloatoffset\fi}{}%
        {\PackageError{paper}{Double-column float spacing patch failed}{}}%
}
\makeatother
\definecolor{baselinecolor}{HTML}{F0F1FD}
\definecolor{robotonlycolor}{HTML}{EDEFF2}
\definecolor{ablationgroupcolor}{HTML}{FFF7E8}
\definecolor{pretrainonlycolor}{HTML}{F1F1F1}

\newcommand{\cmark}{\ding{51}}
\newcommand{\xmark}{\ding{55}}
\newcommand{\sysname}{\textsc{EgoWild2Dex}\xspace}
\newcommand{\geoformer}{\textsc{GeoFormer}\xspace}
\newcommand{\R}{\mathbb{R}}
\newcommand{\glovedata}{glove demonstrations\xspace}

\newcommand{\revo}{Revo2\xspace}
\newcommand{\piper}{Piper\xspace}
\newcommand{\pifive}{$\pi_{0.5}$\xspace}
\newcommand{\pizero}{$\pi_{0}$\xspace}

\def\BibTeX{{\rm B\kern-.05em{\sc i\kern-.025em b}\kern-.08em
    T\kern-.1667em\lower.7ex\hbox{E}\kern-.125emX}}

\begin{document}
\bstctlcite{paper-bib-control}

\title{\LARGE\bfseries \sysname: Learning Dexterous Robotic Manipulation from In-the-Wild Human Experience}

\author{
Kunyang Lin$^{\dagger}$, Xutao Wen, Jingxi Lin, Lanyong Lin, Jiaming Liu,\\[-1pt]
Tianshuo Yang, Xianchi Chen, Yue Han, Yiduo Li, Zhanpeng Zhang, Ping Luo$^{*}$\\[3pt]
{\normalsize The University of Hong Kong \quad Kinetix AI}\\[2pt]
{\color{gray!65}\normalsize
$^{\dagger}$Project lead \quad
$^{*}$Corresponding author}\\[2pt]
{\normalsize\href{https://mmlab.hk/egowild2dex}{\textcolor[rgb]{0.86,0.10,0.45}{\texttt{https://mmlab.hk/egowild2dex}}}}
}

\IEEEaftertitletext{%
    \begin{center}
    \vspace{-4mm}
    \includegraphics[width=\textwidth]{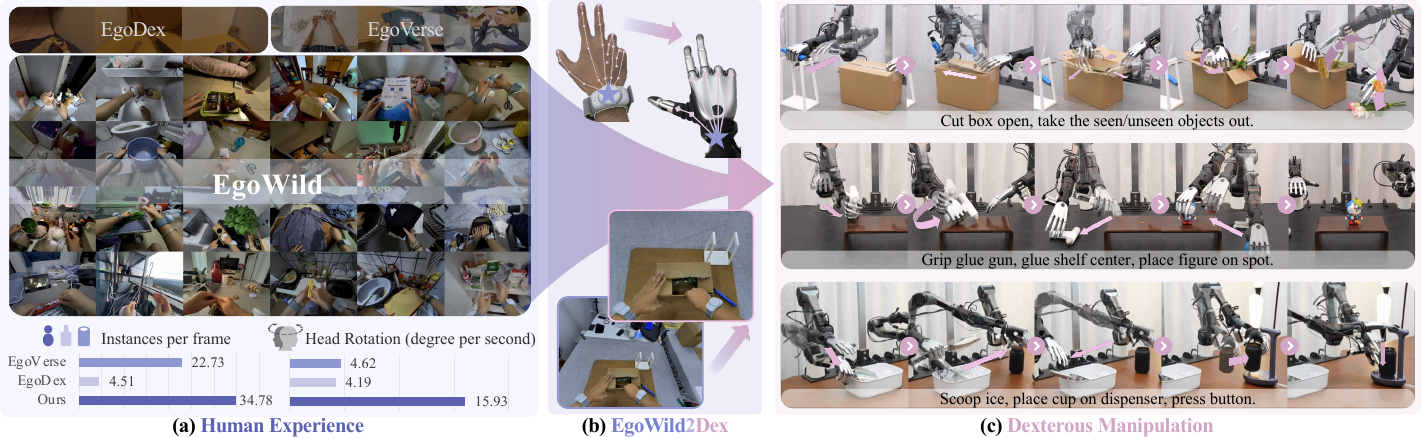}\\[3pt]
    \refstepcounter{figure}\label{fig:teaser}%

        \parbox[t]{\textwidth}{%
        \normalfont\fontsize{8}{9}\selectfont\noindent
        Fig.~\thefigure. \textbf{We introduce \sysname, a framework for learning dexterous robot manipulation from in-the-wild human experience.}
        (a) EgoWild is an in-the-wild egocentric human manipulation dataset captured with head-mounted cameras during everyday activities. On average, each frame of EgoWild contains $34.78$ segmented foreground instances ($7.71\times$ EgoDex~\cite{hoque2025egodex} and $1.53\times$ EgoVerse~\cite{punamiya2026egoverse}), and cumulative head rotation is $15.93^\circ$/s ($3.80\times$ and $3.45\times$, respectively).
        (b) \sysname is a training pipeline that transfers in-the-wild human experience to dexterous robots with limited robot data in a shared robot-native action space. At its core is a differentiable geometric transformer that aligns task-specific ego observations with robot views for co-training.
        (c) We demonstrate our method on three long-horizon bimanual dexterous tasks. Each task presents challenges involving tool use, arm--hand coordination, and contact-rich interaction, and consists of multiple subtasks in sequence.%
    }
    \end{center}
    \vspace{-1mm}
}

\maketitle
\thispagestyle{empty}
\pagestyle{empty}
\flushbottom
\setlength{\parskip}{0pt plus 1pt}
\setlength{\textfloatsep}{10pt plus 2pt minus 2pt}
\setlength{\dbltextfloatsep}{10pt plus 2pt minus 2pt}
\setlength{\floatsep}{8pt plus 2pt minus 2pt}
\setlength{\dblfloatsep}{8pt plus 2pt minus 2pt}
\begin{abstract}
Egocentric human data provide a principled source of supervision for learning dexterous robot manipulation. Unlike prior approaches that often collect such data in constrained or specially constructed environments, we collect in-the-wild egocentric demonstrations in real-world settings, including homes, factories, and pharmacies, \etc, where people perform their ordinary tasks while wearing head-mounted cameras. This collection protocol captures diverse workflows and hand--object interactions across long-tailed object and skill distributions, but also yields visually challenging observations due to scene clutter and head-motion-induced viewpoint changes (a mean cumulative rotation of $\mathbf{15.93^\circ}$/s). To address these issues, we introduce \sysname, which transfers in-the-wild ego-human experience to dual-arm robots with dexterous hands by jointly aligning unstable egocentric views and human motions with robot observations and actions, respectively. This work offers three benefits. First, we introduce \geoformer, a differentiable geometric transformer that warps noisy human observations toward robot observations. Second, we design a human--robot training scheme to bridge the embodiment gap, enabling high task success with limited robot supervision. Third, we release EgoWild, a 538.9-hour in-the-wild egocentric human dataset comprising 179,049 episodes, 125,961 unique task descriptions, and 1,282 object categories. On real robots, \sysname achieves an average success rate of $\mathbf{96.7\%}$ across three long-horizon bimanual dexterous manipulation tasks and an average object-level zero-shot success rate of $\mathbf{33.3\%}$. The data, models, and code will be released.
\end{abstract}

\section{Introduction}
\label{sec:intro}
Large-scale ego-human demonstrations offer a scalable alternative to costly robot teleoperation trajectories. Beyond scale, they capture environmental dynamics, contact-induced object affordances, and dexterous arm--hand kinematics~\cite{ma2026robot,he2024omnih2o,spiritv15}.
Recent works have demonstrated that ego-human data can provide beneficial priors for robot manipulation~\cite{kareer2024egomimic,yang2025egovla,beingh0,egoscale,chen2026robots}. However, existing approaches often rely on massive amounts of non-public data, costly paired human--robot data, or specially constructed scenes~\cite{dyna2026dyna2, beingbeyond2026beingh08, egoscale,hoque2025egodex,beingh0,xie2026human2robot,jain2024vid2robot,wang2023mimicplay}. Moreover, most target gripper-based manipulation~\cite{wang2026humanego,chen2026robots,xie2026human2robot,jain2024vid2robot,wang2023mimicplay,kareer2024egomimic,liu2025egozero} and do not consider multi-finger coordination for long-horizon dexterous tasks.

In this work, we seek to learn from naturally occurring human manipulation in everyday life, where actions are shaped by the available objects, surrounding clutter, and task demands. Such in-the-wild experience exposes the model to diverse objects and hand--object interactions in cluttered scenes with dynamic viewpoints, reducing its reliance on shortcuts specific to staged setups and encouraging it to learn manipulation patterns that transfer across environments.

However, transferring this experience to robots presents two challenges. First, egocentric observations change with both manipulation and head motion. Consequently, the same manipulation can appear different under moving human cameras and fixed robot cameras. Second, broad human experience provides diverse manipulation behaviors but does not directly teach a robot how to execute a specific task under its own embodiment and physical constraints. The challenge is to turn this experience into reliable robot behavior.

To tackle these challenges, we introduce \sysname, a framework for learning dexterous robot manipulation from in-the-wild human experience. We propose Geometric Transformer (\geoformer) for lightweight projective alignment of task-specific ego observations with robot views, eliminating the need for precise camera calibration and the high computational and memory costs of existing 3D projection-and-inpainting pipelines~\cite{shi2026egohumanoid,zhao2026halomi}. It learns a robot-to-ego homography and applies its inverse to bring ego observations closer to the robot view, providing visually aligned human demonstrations for co-training. To turn broad human experience into reliable robot execution, we design a progressive human--robot training scheme. \textbf{Human-to-Robot learning} first learns a broad prior from ego-human demonstrations. \textbf{Human--Robot co-training} then grounds this prior using task-specific human demonstrations and limited robot data. Finally, \textbf{robot-domain refinement} incorporates execution and recovery to reduce the remaining contact and dynamics gap.
All stages share the policy architecture, robot-native action space, and flow-matching objective, while using different data and schedulers.

To support this framework, we construct EgoWild, a large-scale dataset of in-the-wild human manipulation, captured with head-mounted cameras during everyday tasks. As shown in Fig.~\ref{fig:teaser}, EgoWild contains more foreground instances per frame ($34.78$) and greater cumulative head rotation per second ($15.93^\circ$) than EgoDex~\cite{hoque2025egodex} and EgoVerse~\cite{punamiya2026egoverse}, featuring denser clutter and more dynamic viewpoints.

We evaluate our full system on three long-horizon tasks, each comprising multiple subtasks that involve tool use, bimanual coordination, and object manipulation.
Empirical results show that, using less than one hour of robot demonstrations per task, \sysname achieves an average success rate of $96.7\%$ across three long-horizon bimanual dexterous tasks. \sysname also achieves an average object-level zero-shot success rate of $33.3\%$ on unseen objects. These results suggest that combining EgoWild with \geoformer and progressive human--robot training can effectively transfer in-the-wild human experience to robot manipulation.

Collectively, our contributions are threefold: \textbf{(1)} We introduce \geoformer, a differentiable geometric transformer for lightweight ego-to-robot view alignment without explicit 3D reconstruction or inpainting, delivering a speedup of up to $21.9\times$. \textbf{(2)} We develop a progressive human--robot training scheme that transfers broad human manipulation priors to dexterous robots, achieving a $96.7\%$ average success rate across three long-horizon dexterous tasks with less than one hour of robot data per task. \textbf{(3)} We release EgoWild, a 538.9-hour in-the-wild egocentric dataset of unscripted bimanual activities across real-world environments, objects, and workflows. We demonstrate the potential of EgoWild in dexterous manipulation and believe it will benefit research in other domains.

\begin{figure*}[t]
    \centering
    \includegraphics[width=\textwidth]{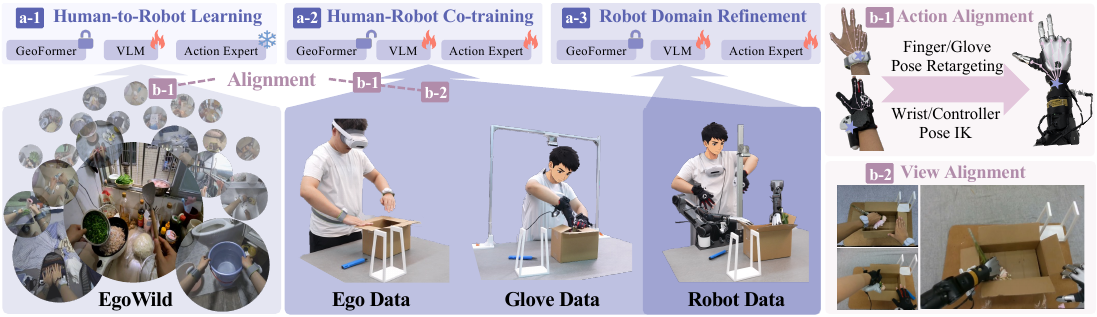}
    \caption{\textbf{Overview of \sysname.}
    (a-1) \textbf{Human-to-Robot learning} builds broad manipulation priors by updating the vision--language model (VLM) on \textbf{538.9 hours} of EgoWild data.
    (a-2) \textbf{Human--Robot co-training} first trains \geoformer on unpaired task-specific ego and robot images, then freezes it to align ego observations while updating the VLM and action expert. The task-specific ego, glove, and robot data used in this stage have a combined duration equivalent to \textbf{just 1.6\%} of the data used in Stage~1.
    (a-3) \textbf{Robot-Domain refinement} improves execution and recovery by updating the VLM and action expert on robot data totaling \textbf{just 0.1\%} of Stage~1's data duration.
    (b-1) Inverse kinematics (IK) and hand retargeting map all demonstrations into one robot-native action space. (b-2) \geoformer aligns ego observations to the robot view, while glove and robot data share the same camera configuration.}
    \label{fig:pipeline}
\end{figure*}

\section{Related Work}
\label{sec:related}

\subsection{Vision--Language--Action Models for Dexterous Hands}
Vision--language--action (VLA) models predict robot actions from visual observations and language instructions. Existing VLAs acquire broad manipulation capabilities from Internet-scale knowledge and heterogeneous robot data~\cite{rt2,openvla,pi0,pi05,gr3,gr00tn1}, but are predominantly developed around arm--gripper action spaces that do not specifically capture fine-grained multi-finger coordination. Unlike grippers, multi-DoF hands require coordinated control of multiple motors, and the resulting commands can exhibit jitter and abrupt changes under limited supervision. To study this, hierarchical dexterous systems connect semantic planners to grasping or reusable low-level skills~\cite{dexgraspvla,being0,dexskills}; although these systems are modular, this separation can weaken closed-loop adaptation when contact outcomes deviate from the planned skill. End-to-end dexterous VLAs directly model high-DoF arm--hand actions~\cite{wen2025gr,dexora,beingh0,wei2025cyclemanip}, but demand substantial robot-domain supervision and remain vulnerable to compounding errors over long horizons, motivating additional on-policy or reinforcement-learning post-training~\cite{pistar06,grrl}. In contrast, \sysname transfers in-the-wild human experience to dexterous VLAs through geometric view alignment and progressive human--robot training, reducing reliance on robot demonstrations.

\subsection{Ego-Human Data for Dexterous Robot Learning}
Early work primarily uses human data either to pretrain visual representations or to support high-level planning, leaving downstream action learning to robot supervision~\cite{nair2022r3m,xiao2022mvp,lin2023learning,zheng2025flare,wang2023mimicplay,xu2023xskill,xu2024flow}. More recent methods exploit egocentric sensing and hand tracking as dense action supervision through explicit human--robot alignment and co-training~\cite{kareer2024egomimic,qiu2025humanoid,tao2025dexwild}, pretraining followed by inverse kinematics and retargeting~\cite{yang2025egovla}, or large-scale cross-embodiment pretraining~\cite{kareer2025emergence,beingh0,egoscale}. In parallel, new datasets expand task, object, and environment coverage~\cite{hoque2025egodex,li2026egolive,li2026open,punamiya2026egoverse}, while some methods seek to synthesize robot-domain observations~\cite{li2026ace,shi2026egohumanoid,wang2026ego2robot}. Rather than scaling data indiscriminately, \sysname pairs a natural, manipulation-rich ego dataset with an explicit action- and view-alignment recipe. This data-efficient design achieves strong long-horizon bimanual dexterity with under one hour of robot data per task.

\section{Methodology}
\label{sec:method}

\looseness=-1 To transfer in-the-wild human experience to dexterous robots, we propose \sysname, a framework combining geometric view alignment with progressive human--robot training (Fig.~\ref{fig:pipeline}). Specifically, \geoformer aligns task-specific ego observations with the robot view, providing visually aligned human demonstrations for co-training. Using a shared action space throughout, the training scheme first learns broad manipulation priors, then adapts them through task-specific human--robot co-training, and finally refines robot execution with expert demonstrations and recovery trajectories.  Algorithm~\ref{alg:pipeline} summarizes the training and deployment procedure.

\subsection{Geometric Transformer}
\label{sec:geoformer}
\begin{figure}[!t]
    \centering
    \includegraphics[width=\columnwidth]{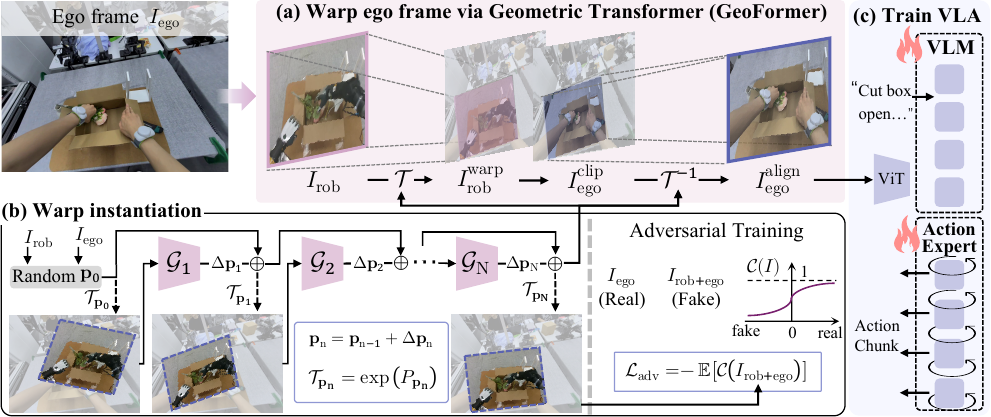}
    \caption{\textbf{Geometric Transformer (\geoformer).}
    (a) Alignment path: \geoformer predicts a robot-to-ego warp $\mathcal{T}$ and applies $\mathcal{T}^{-1}$ to align ego observations with the robot view. (b) Warp module: predictors $\{\mathcal{G}_n\}$ refine homography parameters to make warped robot images appear natural when composited onto ego images, under adversarial supervision from the critic $\mathcal{C}$. (c) VLA training: the aligned ego images are used alongside robot observations to train the VLA policy through human--robot co-training.}
    \label{fig:geoformer}
\end{figure}

\looseness=-1 A key challenge of using egocentric human data is that the same action appears different under head-mounted human cameras and stationary robot cameras. Traditional methods use 3D projection followed by inpainting to align the viewpoints~\cite{shi2026egohumanoid,zhao2026halomi}. These methods require precise camera calibration and are computationally expensive. To overcome this, as shown in Fig.~\ref{fig:geoformer}, we propose \geoformer, aligning task-specific ego observations to the robot view using a lightweight
projective warp. \geoformer learns from
unpaired images $I_{\mathrm{ego}},I_{\mathrm{rob}}\in\mathbb{R}^{3\times H_{\mathrm{img}}\times W_{\mathrm{img}}}$,
where $H_{\mathrm{img}}$ and $W_{\mathrm{img}}$ denote image height and width. It
randomly pairs each ego image with a robot-view image sampled from the
corresponding task's robot-image pool during training. The model
represents their geometric difference using an eight-dimensional homography parameter vector $\mathbf{p}=[p_1,\ldots,p_8]^\top$, which is then mapped to a homography $\mathcal T$ through
\begin{equation}
\begin{aligned}
    P_{\mathbf p}&:=
    \begin{bmatrix}
        p_3 & p_2 & p_1 \\
        p_6 & -p_3-p_7 & p_5 \\
        p_4 & p_8 & p_7
    \end{bmatrix},\\
    \mathcal{T}_{\mathbf p}&:=\exp(P_{\mathbf p})=\lim_{K\to \infty}\sum_{k=0}^{K}\frac{P_{\mathbf p}^k}{k!},
\end{aligned}
\end{equation}
where $P_{\mathbf p}$ is a traceless matrix of eight projective
parameters, and its exponential $\mathcal{T}_{\mathbf p}$ defines
the homography.

Starting from a perturbed estimate $\mathbf{p}_0$, \geoformer progressively refines the
warp. At the $n$-th step, it warps the robot image using the current estimate,
\begin{equation}
    I_{\mathrm{rob},n-1}^{\mathrm{warp}}=\operatorname{Warp}(I_{\mathrm{rob}};\mathcal{T}_{\mathbf p_{n-1}}),
\end{equation}
where $\operatorname{Warp}(I;\mathcal{T})$ denotes warping image $I$ by the homography
$\mathcal{T}$, and $I_{\mathrm{rob},n-1}^{\mathrm{warp}}$ is the robot image warped
toward the ego view using the parameter estimate $\mathbf p_{n-1}$ from the
previous step. The next residual is predicted as
\begin{equation}
    \Delta \mathbf p_n=\mathcal G_n(I_{\mathrm{ego}},I_{\mathrm{rob},n-1}^{\mathrm{warp}}),
\end{equation}
where $\mathcal G_n$ is the $n$-th predictor that takes the concatenation of
the ego image and the currently warped robot image as input and outputs an
eight-dimensional residual in homography parameter space. Each predictor uses a convolutional neural network (CNN) to extract visual features, followed by a multilayer perceptron (MLP) to regress the residual~\cite{jaderberg2015spatial,lin2018stgan}.
We then add this residual to the previous estimate, yielding $\mathbf p_n=\mathbf p_{n-1}+\Delta \mathbf p_n$.

The final robot-to-ego warp is
$\mathcal{T}=\exp(P_{\mathbf p_N})$, where $N$ is the total number of cascaded refinement steps.
During training, as illustrated in Fig.~\ref{fig:geoformer}, we apply
$\mathcal{T}$ to both the robot image and an all-ones mask of the same dimensions to obtain
\begin{equation*}
    I_{\mathrm{rob}}^{\mathrm{warp}} = \operatorname{Warp}(I_{\mathrm{rob}};\mathcal{T}),\quad
    M_{\mathcal{T}} = \operatorname{Warp}(\mathbf{1};\mathcal{T}),
\end{equation*}
where $\mathbf{1}$ is an all-ones image with the same dimensions as
$I_{\mathrm{rob}}$, and out-of-bounds samples are zero-padded.
The resulting mask selects valid warped content to form the composite
\begin{equation}
    I_{\mathrm{rob+ego}}=M_{\mathcal{T}}\odot I_{\mathrm{rob}}^{\mathrm{warp}}+(1-M_{\mathcal{T}})\odot I_{\mathrm{ego}},
\end{equation}
where $\odot$ denotes element-wise multiplication and $M_{\mathcal{T}}$
selects valid warped robot content, retaining the ego background elsewhere.
The geometric predictor is trained with adversarial loss to warp robot-view
content and paste it into the ego image as a natural-looking composite. It
learns to fool a Wasserstein critic~\cite{gulrajani2017improved} that is trained simultaneously to distinguish real
ego images from predictor-generated composites. Geometric regularization further keeps the pasted region at a plausible scale and maintains sufficient overlap
with the ego canvas, preventing tiny patches or excessive out-of-frame warps.

After training \geoformer, the critic is discarded and the predictor is frozen during subsequent human--robot policy co-training.
The learned warp maps robot views to ego views; policy inputs therefore
use its inverse:
\begin{equation}
    {I}_{\mathrm{ego}}^{\text{align}}
    =\operatorname{Warp}(M_{\mathcal{T}}\odot I_{\mathrm{ego}};
    \mathcal{T}^{-1}),
\end{equation}
where $I_{\mathrm{ego}}^{\text{align}}$ is the ego observation aligned toward
the robot view by the inverse homography $\mathcal{T}^{-1}$. Despite fixed-resolution inputs, \geoformer supports inverse warping at arbitrary resolutions with coordinate scaling.
The aligned frame is then processed by the same VLA vision encoder as a robot image. Compared with the Project+Inpaint pipeline using MoGe-based 3D reprojection~\cite{wang2025moge} with LaMa inpainting~\cite{suvorov2022lama}, \geoformer achieves a $21.9\times$ speedup and improves both robot-view image similarity and downstream robot task performance (Sec.~\ref{sec:visual-alignment-exp}).

\subsection{Progressive Human--Robot Training}
\label{sec:model_architecture}
As shown in Fig.~\ref{fig:pipeline}, we use a VLA architecture comprising a vision--language model backbone that processes images and task instructions, and a flow-matching action expert that predicts robot-native action chunks. We leverage a consistent action space to learn across human and robot demonstrations.

\paragraph{Shared robot-native action space}

We calibrate operators' arm poses to approximately match the robot's reference configuration, establishing a common spatial reference for motion retargeting. Subsequent wrist and finger movements are converted into robot-native commands relative to this reference, reducing mismatches caused by arbitrary initial poses and providing consistent action supervision across training stages. The shared action vector contains the arm joint targets and hand motor commands for both sides,
\begin{equation}
A_t
:=
\begin{bmatrix}
\mathbf{a}^{\mathrm{arm}}_{L,t} \\
\mathbf{a}^{\mathrm{hand}}_{L,t} \\
\mathbf{a}^{\mathrm{arm}}_{R,t} \\
\mathbf{a}^{\mathrm{hand}}_{R,t}
\end{bmatrix}
\in \mathbb{R}^{d_a},
\qquad
d_a = 2\left(d_{\mathrm{arm}} + d_{\mathrm{hand}}\right),
\label{eq:action-space-formal}
\end{equation}
where $\mathbf{a}^{\mathrm{arm}}_{j,t}$ and $\mathbf{a}^{\mathrm{hand}}_{j,t}$ are the commanded arm joint angles and hand motor values for side $j\in\{L,R\}$, and $d_{\mathrm{arm}}$ and $d_{\mathrm{hand}}$ denote the dimensions of the arm-joint and hand-motor command vectors for each side. Calibrated wrist and finger motions from every source are converted through the same retargeting and IK pipeline used for robot teleoperation. Across all stages, the robot-native action chunks serve as targets for the same flow-matching loss~\cite{pi0,pi05}, keeping the training objective and output representation unchanged.

Shared action targets make human and robot demonstrations compatible for optimization, but their contributions to learning remain complementary: broad human experience provides reusable manipulation priors, task-specific demonstrations connect these priors to the target task, and robot execution supplies contact and recovery experience. Motivated by this observation, we structure training around the progression, moving from broad prior learning through human--robot co-training to robot-domain refinement while retaining the same policy architecture.

\paragraph{Human-to-Robot learning}
The first stage learns broad bimanual semantics and motion patterns from EgoWild demonstrations. Ego-human observations are preserved in their original form, allowing the policy to learn from the visual and behavioral diversity of the in-the-wild corpus.

\begin{algorithm}[t!]
\small
\caption{Training and deployment pipeline}
\label{alg:pipeline}
\begin{algorithmic}[1]
\REQUIRE In-the-wild ego data $\mathcal{D}_{\mathrm{HB}}$, task-specific ego data $\mathcal{D}_{\mathrm{HT}}$, \glovedata $\mathcal{D}_{\mathrm{G}}$, robot data $\mathcal{D}_{\mathrm{R}}$, and instruction $\ell$.
\ENSURE A deployed bimanual dexterous VLA policy $\pi$.
\STATE \textbf{Build shared robot-native supervision:}
\STATE Initialize processed subsets $\widetilde{\mathcal{D}}_{s}$ for $s\in\{\mathrm{HB},\mathrm{HT},\mathrm{G},\mathrm{R}\}$.
\FOR{each source $s$ and trajectory $\tau\in\mathcal{D}_{s}$}
    \STATE Read wrist poses, finger signals, images, and instructions.
    \STATE Apply shared calibration, dual-arm IK, and hand retargeting to obtain commands $A_t\in\R^{d_a}$.
    \STATE Preserve source observations and store $(o_t,\ell,\mathbf q_t,A_t)$ in $\widetilde{\mathcal{D}}_{s}$.
\ENDFOR
\STATE \textbf{Progressively train the VLA policy:}
\STATE Perform Human-to-Robot learning on $\widetilde{\mathcal{D}}_{\mathrm{HB}}$.
\STATE Begin Human--Robot co-training by training \geoformer on unpaired images from $\mathcal{D}_{\mathrm{HT}}$ and $\mathcal{D}_{\mathrm{R}}$.
\STATE Freeze \geoformer and use it to align ego observations in $\widetilde{\mathcal{D}}_{\mathrm{HT}}$.
\STATE Jointly train the VLM and action expert on $\widetilde{\mathcal{D}}_{\mathrm{HT}}\cup\widetilde{\mathcal{D}}_{\mathrm{G}}\cup\widetilde{\mathcal{D}}_{\mathrm{R}}$.
\STATE Refine in the robot domain using expert execution and recovery trajectories from $\widetilde{\mathcal{D}}_{\mathrm{R}}$.
\STATE \textbf{Deploy with online correction:}
\WHILE{executing a task on the real robot}
    \STATE Predict and execute action chunks from $\pi(\tilde{o}_t,\ell,\mathbf{q}_t)$.
    \IF{human correction is required}
        \STATE Apply Anchored Delta-Cmd and add corrections to $\widetilde{\mathcal{D}}_{\mathrm{R}}$, then continue refinement.
    \ENDIF
\ENDWHILE
\RETURN Deployed policy $\pi$.
\end{algorithmic}
\end{algorithm}

\paragraph{Human--Robot co-training}
The second stage first trains \geoformer on unpaired task-specific ego and robot images using the view-alignment objective in Sec.~\ref{sec:geoformer}. Then, we freeze \geoformer and use it to align task-specific ego observations while jointly training the VLM and action expert on the aligned ego data, robot-view \glovedata, and a small amount of robot demonstration data. This stage aims to adapt the broad policy to target tasks.

\paragraph{Robot-domain refinement}
The final stage uses expert robot demonstrations and recovery trajectories to reduce the remaining embodiment and contact-dynamics gaps. Recovery trajectories are collected using Dataset Aggregation (DAgger)~\cite{kelly2019hgdagger}, which augments the training set with expert recoveries from policy-visited states.

Every stage optimizes the same conditional mapping $(\tilde{o}^{s}_{t},\ell,\mathbf{q}_t)\mapsto\mathbf{A}_{t}$, while the policy parameters continue to be updated. Here, $\tilde{o}^{s}_{t}$ is the visual observation at time $t$ from data source $s$, after \geoformer alignment when applicable; $\mathbf{q}_t\in\R^{d_a}$ is the current measured joint state; $\ell$ is the language task instruction; and $\mathbf{A}_{t}:=[A_t,A_{t+1},\ldots,A_{t+H-1}]\in\R^{d_a\times H}$ is an action chunk of horizon $H$. Each $A_t$ contains the arm joint targets and hand motor commands defined in Eq.~\eqref{eq:action-space-formal}.

\subsection{Data Sources}
\label{sec:dataset-construction}

\paragraph{Ego-Human Demonstrations (EgoWild)}
We create EgoWild by collecting bimanual human behavior with VR devices at physically distinct real-world sites, including homes, supermarkets, parcel stations, factories, pharmacies, and offices, rather than recreating a small set of spaces. Objects, clutter, lighting, and layouts remain in their native state, exposing natural environmental diversity and scene clutter. Collection is open-ended and unscripted: operators plan and complete real tasks according to the local context without predefined action sequences. Raw long recordings preserve the behavioral chain, including observation, decision making, task switching, trial and error, locomotion, and inspection. For Stage~1 training, we retain quality-filtered manipulation clips with limited translational movement, keeping the supervision focused on manipulation.

EgoWild combines high-resolution egocentric video, tracked hand motion, and temporally structured language annotations. Videos are captured at $2048\!\times\!1536$, compared with $1920\!\times\!1080$ for EgoDex~\cite{hoque2025egodex} and $640\!\times\!360$ for EgoVerse~\cite{punamiya2026egoverse}. Collectors wear a calibrated tracker on each hand: in our 3D reaching test, median absolute fingertip-position error decreases from $4.21$~cm without the tracker to $0.66$~cm with it. Language annotations are organized at three timestamped levels: atomic-action segments, segment descriptions, and complete skills composed of multiple segments. In contrast, EgoDex primarily provides episode-level descriptions, while EgoVerse provides episode subtasks without corresponding temporal boundaries. The final processed corpus used for Stage~1 training contains $179{,}049$ episodes and $538.9$ hours of egocentric activity ($58.2$M frames at $30$\,Hz). As shown in Fig.~\ref{fig:teaser}, we use SAM~3.1~\cite{carion2025sam3} to segment foreground instances to quantify visual complexity. EgoWild contains an average of $34.78$ foreground instances per frame, compared with $4.51$ in EgoDex and $22.73$ in EgoVerse. It also exhibits substantially greater mean cumulative head rotation per second ($15.93^\circ$ versus $4.19^\circ$ and $4.62^\circ$). These metrics reflect denser scene clutter and more dynamic viewpoints in EgoWild.
Additionally, we collect a task-specific corpus following the same data-collection pipeline in recreated downstream layouts for adaptation. \geoformer then transforms its ego observations toward the robot viewpoint; see Sec.~\ref{sec:geoformer}. For motion alignment, we express all poses in the initial headset coordinate frame and calibrate the VR pose corresponding to the robot-arm zero configuration. We then compute hand motions relative to this reference and obtain robot-relative trajectories through a fixed coordinate transform.

\paragraph{Glove Demonstrations}
Glove demonstrations provide robot-view observations and retargeted arm--hand action labels. As depicted in Fig.~\ref{fig:pipeline}, the operator performs downstream tasks under the same three-camera setup used at deployment, preserving viewpoint, workspace scale, and occlusion patterns of the robot. VR controllers track arm end-effector poses, while data gloves capture finger articulation; relative-pose mapping, dual-arm IK, and hand retargeting convert these signals into robot-native commands. Compared with headset-based hand tracking, controllers and gloves provide more accurate wrist-pose and finger-articulation measurements, respectively. Both devices are the same as those used for robot teleoperation; combined with the shared action mapping, this is expected to bring the resulting action supervision closer to the distribution of commands received by the robot. Thus, \glovedata supply deployment-matched observations and higher-fidelity action supervision before costly physical execution.

\paragraph{Real-Robot Demonstrations}
Real-robot teleoperation applies the same mapping online and sends the resulting commands to the physical robot. Built on XRoboToolkit~\cite{zhao2026xrobotoolkit}, the system supports bare-hand tracking, auxiliary trackers, and VR controllers; we use VR controllers as they provide the most stable pose signal in our setup. Teleoperation supplies contact-grounded demonstrations and subsequently supports human intervention; light low-pass filtering suppresses sensing noise in the hand commands.

\subsection{Reliable Deployment and Recovery Collection}
\label{sec:deployment}
Contact-rich dexterous manipulation requires robust execution. To this end, we develop a reliable system for teleoperation, policy deployment, and human correction: we adopt curated calibration and filtering to produce smooth action targets; for low-level control, we apply compliant torque control to preserve contact intent while absorbing small pose errors; and overlap blending prevents discontinuities between predicted action chunks. For recovery collection, we propose Anchored Delta-Cmd, a human-takeover mechanism that anchors robot commands at their takeover values and updates them using subsequent relative operator motion, avoiding command jumps. These engineering components make data collection and closed-loop deployment robust and serve as the foundational infrastructure for our whole learning system.

\section{Experiments}
\label{sec:experiments}

We evaluate \sysname on real-world bimanual dexterous manipulation through four questions:
\textbf{RQ1:} How does \sysname compare with direct robot-only adaptation under limited robot data?
\textbf{RQ2:} How does each component contribute within the progressive training pipeline?
\textbf{RQ3:} How does \geoformer affect visual transfer quality and efficiency?
\textbf{RQ4:} How does the amount of human data affect downstream task performance, and how well does the learned policy generalize and transfer across objects and embodiments?

\subsection{Experimental Setup}
\label{sec:exp-setup}

\paragraph{Robot platform}
We conduct experiments mainly on two six-joint \piper arms equipped with
two six-channel \revo hands. Observations come from a fixed top-down camera
and wrist cameras that move with the arms. The robot uses the arm--hand
commands defined in Eq.~\eqref{eq:action-space-formal}.

\paragraph{Evaluation tasks}
Our three long-horizon bimanual tasks comprise sequential subtasks requiring dexterous arm--hand coordination and contact-rich interaction. Scores measure subtask completion; numbers in parentheses below indicate the points awarded for completing each subtask.

\textbf{Task I: Open-Box.}
The robot is required to grasp a utility knife ($2$), cut the tape ($2$), open the box ($4$), and remove the bouquet ($2$).

\textbf{Task II: Glue-Figure.}
The robot grasps a hot glue gun ($2$), applies glue ($3$: dispensing $1$, accurate placement $2$), picks up the figure ($2$), and places it on the stand ($3$: placement $1$, alignment $2$).

\textbf{Task III: Ice-Water.}
The robot grasps the scoop ($2$), transfers ice into a cup ($3$: scooping $1$, pouring $2$), positions the cup on the dispenser ($2$), and adds water ($3$: lever contact $1$, aligned dispensing $2$).

\paragraph{Metrics}
We report the mean $10$-point completion score and success rate over $10$ independent trials per method and task; success requires completing the full instruction.

\subsection{Motivating Study: Limits of Robot-Only Adaptation}
\label{sec:exp-baselines}
\label{sec:main-results}

\looseness=-1 We compare direct robot-only adaptation of \pizero~\cite{pi0}, \pifive~\cite{pi05}, and \textsc{GR00T-N1.7} (pretrained for dexterous hand manipulation on over $20{,}000$ hours of human egocentric video~\cite{nvidia2026groot17release}) against the complete \sysname pipeline. Each model is fine-tuned on the same $100$ expert robot demonstrations per task.

Table~\ref{tab:main-results} evaluates direct adaptation of public pretrained VLAs to long-horizon bimanual dexterous manipulation using only $100$ expert robot demonstrations per task. The full \sysname pipeline serves as an end-to-end reference.

\begin{table}[t!]
\centering
\caption{\textbf{Direct adaptation of public pretrained VLAs.}
We train the public pretrained VLAs on limited robot expert data and compare their performance with that of our full pipeline.}
\label{tab:main-results}
\resizebox{\columnwidth}{!}{%
\begin{tabular}{l|cccc}
\toprule
Method & Open-Box & Glue-Fig. & Ice-Water & Avg. \\
\midrule
\multicolumn{5}{@{}l@{}}{\raisebox{0.6ex}{\emph{Direct fine-tuning of pretrained VLAs}}} \\[-0.2ex]
\pizero~\cite{pi0}
& 2.0/0.0 & 3.7/0.0 & 2.4/10.0 & 2.7/3.3 \\
\pifive~\cite{pi05}
& 4.0/0.0 & 4.1/10.0 & 4.6/10.0 & 4.2/6.7 \\
\textsc{GR00T-N1.7}~\cite{nvidia2026groot17release}
& 2.5/0.0 & 1.4/0.0 & 2.4/0.0 & 2.1/0.0 \\
\midrule
\multicolumn{5}{@{}l@{}}{\raisebox{0.6ex}{\emph{Human-robot alignment and co-training (full pipeline)}}} \\[-0.2ex]
\cellcolor{baselinecolor}\textbf{\sysname (Ours)}
& \cellcolor{baselinecolor}\textbf{9.5/90.0}
& \cellcolor{baselinecolor}\textbf{10.0/100.0}
& \cellcolor{baselinecolor}\textbf{10.0/100.0}
& \cellcolor{baselinecolor}\textbf{9.8/96.7} \\
\specialrule{\heavyrulewidth}{0pt}{0pt}
\end{tabular}
}

    \vspace{4pt}
    \caption{\textbf{Ablation of the progressive training pipeline.}
    We ablate the two preceding human-data stages. The superscript $\ddagger$ denotes robot-domain refinement augmented with DAgger.}
    \label{tab:stage_ablation}

    \resizebox{0.82\columnwidth}{!}{%
    \begin{tabular}{ccccc}
        \toprule
        \multicolumn{3}{c}{Training Recipe}
        & \multicolumn{2}{c}{Performance} \\
        \cmidrule(lr){1-3}\cmidrule(lr){4-5}
        \shortstack{Human-to-Robot\\learning}
        & \shortstack{Human--Robot\\co-training}
        & \shortstack{Robot-domain\\refinement}
        & \raisebox{0.9ex}{Score}
        & \raisebox{0.9ex}{Succ.} \\
        \midrule
        \xmark & \xmark & \cmark & 1.4 & 0.0\% \\
        \cmark & \xmark & \cmark & 5.1 & 20.0\% \\
        \xmark & \cmark & \cmark & 7.5 & 40.0\% \\
        \cmark & \cmark & \cmark & 8.8 & 70.0\% \\
        \specialrule{\lightrulewidth}{\aboverulesep}{0pt}
        \cellcolor{baselinecolor}\cmark
        & \cellcolor{baselinecolor}\cmark
        & \cellcolor{baselinecolor}\cmark$^{\ddagger}$
        & \cellcolor{baselinecolor}\textbf{9.5}
        & \cellcolor{baselinecolor}\textbf{90.0\%} \\
        \specialrule{\heavyrulewidth}{0pt}{0pt}
    \end{tabular}
    }
\end{table}

\begin{figure}[t!]
    \centering
    \includegraphics[width=\columnwidth]{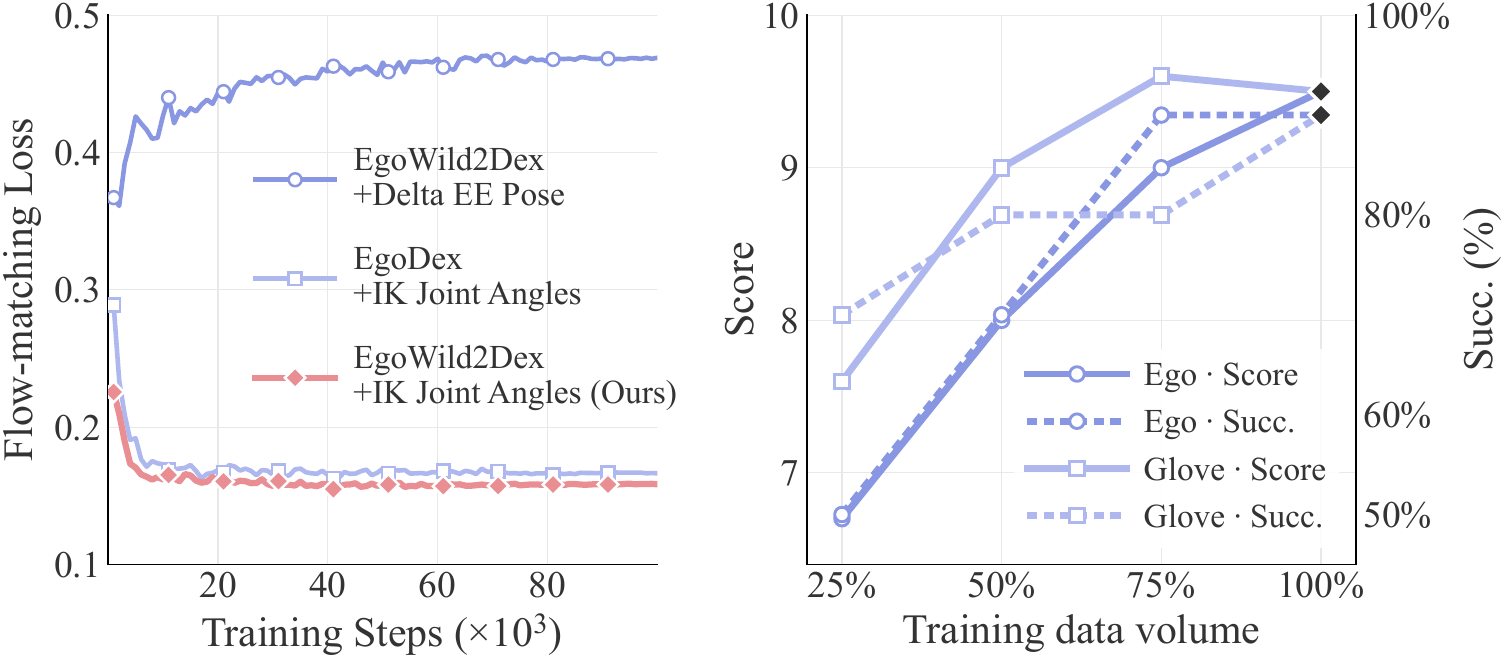}
    \begin{tabular*}{0.8\columnwidth}{@{\extracolsep{\fill}}cc@{}}
        \scriptsize (a) Stage-1 evaluation loss &
        \scriptsize \quad(b) Human--Robot co-training data scaling
    \end{tabular*}%
    \caption{\textbf{Human-to-Robot learning and human-data scaling.}
    (a) Evaluation loss across action representations and ego-human corpora. (b) Independently varying task-specific ego data and \glovedata while keeping the remaining Human--Robot co-training sources fixed.}
    \label{fig:pretraining-and-scaling}
\end{figure}

\begin{figure}[t!]
    \centering
    \includegraphics[width=\columnwidth]{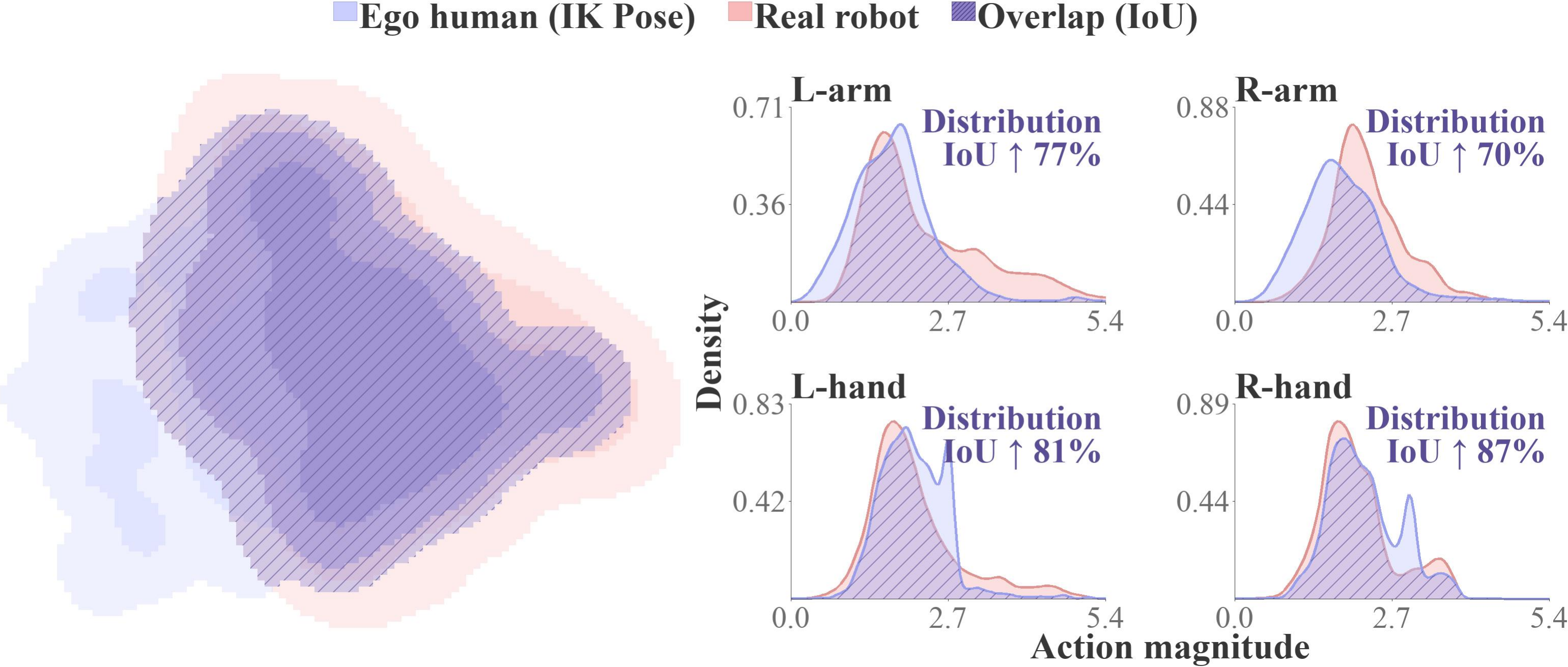}
    \begin{tabular*}{0.84\columnwidth}{@{\extracolsep{\fill}}cc@{}}
        \scriptsize (a) Pooled PCA support &
        \scriptsize (b) Group-wise magnitudes
    \end{tabular*}
    \caption{\textbf{Distributional alignment between retargeted ego-human and real-robot actions.}
    (a) Occupied support in the first two principal components of pooled normalized actions.
    (b) KDEs of normalized action magnitude for the left/right arms and hands.}
    \label{fig:action-distribution-alignment}
\end{figure}

Direct robot-only adaptation remains difficult despite initialization from large public pretrained models. The best adapted checkpoint, \pifive, reaches an average completion score of $4.2$ but completes only $6.7\%$ of trials end-to-end, showing the limitations of direct adaptation with limited robot supervision and motivating a more effective adaptation strategy. Increasing the number of fine-tuning steps further lowers the training loss but brings little performance gain, reaching only $5.3/0.0\%$ on Task~I. For reference, the complete \sysname pipeline reaches a $96.7\%$ average success rate. \sysname incorporates Human-to-Robot learning from in-the-wild ego-human data, task-aligned ego and glove demonstrations, and robot-domain refinement. The contribution of these components is examined next through ablations. To enable rapid iteration while controlling real-robot evaluation time and cost, subsequent ablations focus on the comparatively challenging Open-Box task (Task~I).

\subsection{Ablation of Progressive Training}
\label{sec:ablation}

In this section, we examine the contributions of human-to-robot learning, human–robot co-training, and recovery-augmented robot-domain refinement. Human--robot co-training uses 1,000 task-specific ego demonstrations, 1,000 glove demonstrations, and 100 expert robot demonstrations per task. For robot-domain refinement, we first train on the same 100 expert robot demonstrations used in Human--Robot co-training. We then deploy the policy to collect 10 short DAgger recovery segments and continue fine-tuning on the combined robot data.

Ablation results in Table~\ref{tab:stage_ablation} highlight the criticality of each component.
Conducting robot-domain refinement alone, without either human-data learning stage, achieves only $1.4/0.0\%$. Adding only Human-to-Robot learning increases the performance to $5.1/20.0\%$, while adding only Human--Robot co-training reaches $7.5/40.0\%$, indicating that task-aligned human supervision provides stronger direct grounding. Combining both stages raises performance to $8.8/70.0\%$. This confirms our hypothesis that the broad prior and task-specific grounding are complementary. Anchored Delta-Cmd DAgger further improves performance to $9.5/90.0\%$ through policy-failure recoveries that address the execution-distribution gap.

\subsection{Analysis of Human-to-Robot Learning}
\label{sec:pretraining-ablation}

We investigate how action representation and the choice of ego-human dataset affect Stage~1 learning. We also compare retargeted human and real-robot command distributions to assess their statistical similarity in the shared action space.

We compare delta end-effector poses with absolute robot-native arm and hand commands on EgoWild. In Fig.~\ref{fig:pretraining-and-scaling}(a), evaluation loss decreases and converges with absolute joint-angle targets but increases with delta end-effector pose targets. These results validate our choice of absolute robot-native commands for Human-to-Robot learning.

To assess this action space, we compare retargeted human commands with commands from real-robot demonstrations in Fig.~\ref{fig:action-distribution-alignment}. We measure overlap in a shared PCA projection and similarity between normalized action-magnitude distributions for each arm and hand.
From $10{,}000$ randomly sampled training frames in each of the ego-human and robot domains, we obtain a PCA-support IoU of $0.73$, while the left-arm, left-hand, right-arm, and right-hand density overlaps are $0.77$, $0.81$, $0.70$, and $0.87$, respectively (mean $0.79$). These results suggest our retargeting pipeline aligns human actions with the robot-native action space.

We next compare EgoWild with EgoDex~\cite{hoque2025egodex}, processing both datasets with the same IK and hand-retargeting pipeline and using the same absolute robot-native action representation. EgoDex contains $314{,}834$ episodes, $1.76\times$ as many as EgoWild ($179{,}049$ episodes). Nevertheless, the model trained on EgoWild achieves a lower evaluation loss in Fig.~\ref{fig:pretraining-and-scaling}(a). We attribute this advantage to EgoWild's variation in environments, objects, viewpoints, and execution styles, which discourages setup-specific shortcuts and promotes manipulation patterns that transfer across environments.

\subsection{Efficiency and Quality of \geoformer}
\label{sec:visual-alignment-exp}

To evaluate the efficiency and quality of \geoformer, we compare \geoformer-aligned input with raw ego input and Project+Inpaint-aligned input~\cite{wang2025moge,suvorov2022lama}.

\begin{figure}[t!]
    \centering
    \setlength{\tabcolsep}{0pt}
    \begin{tabular}{@{}*{4}{>{\centering\arraybackslash}p{0.25\columnwidth}}@{}}
        \includegraphics[width=0.94\linewidth]{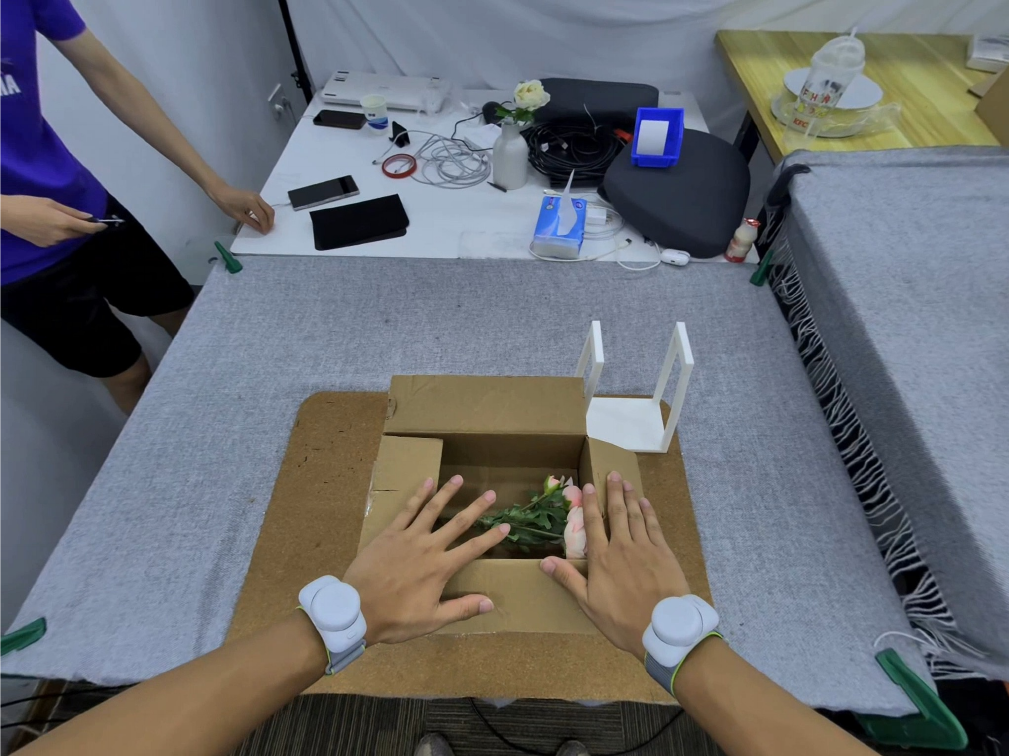} &
        \includegraphics[width=0.94\linewidth]{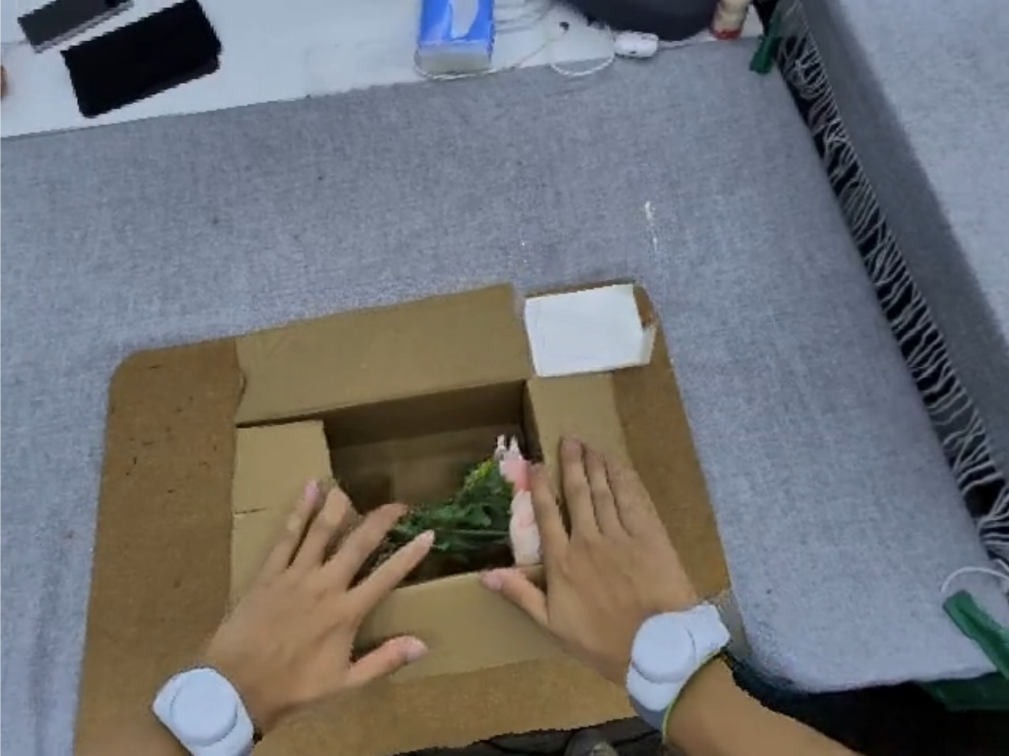} &
        \includegraphics[width=0.94\linewidth]{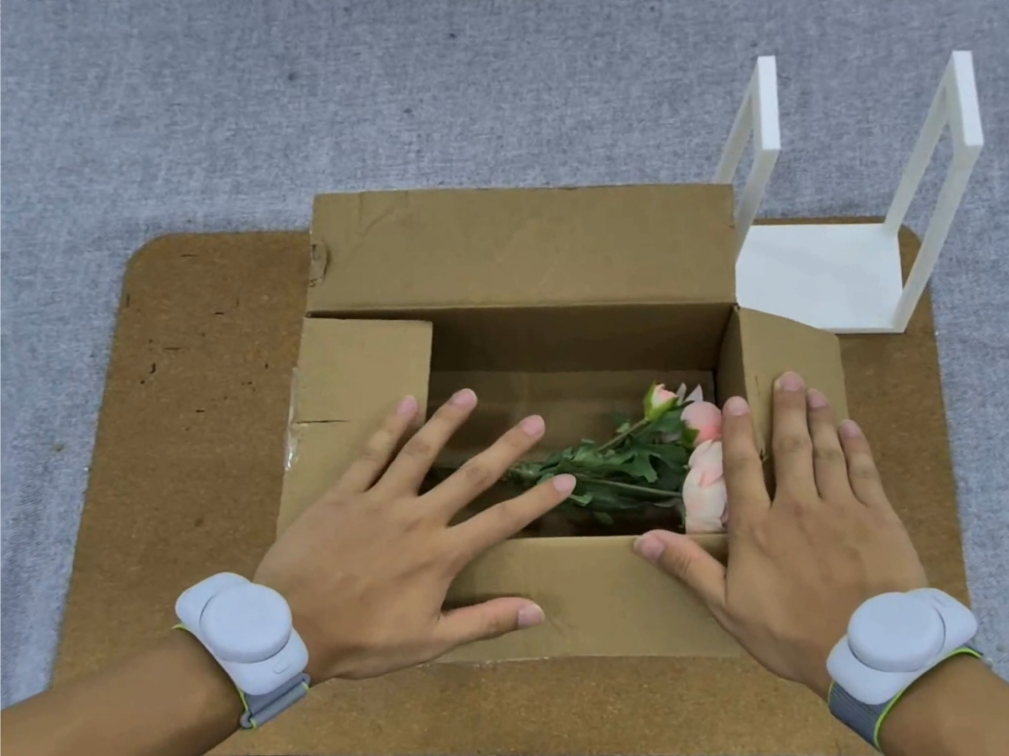} &
        \includegraphics[width=0.94\linewidth]{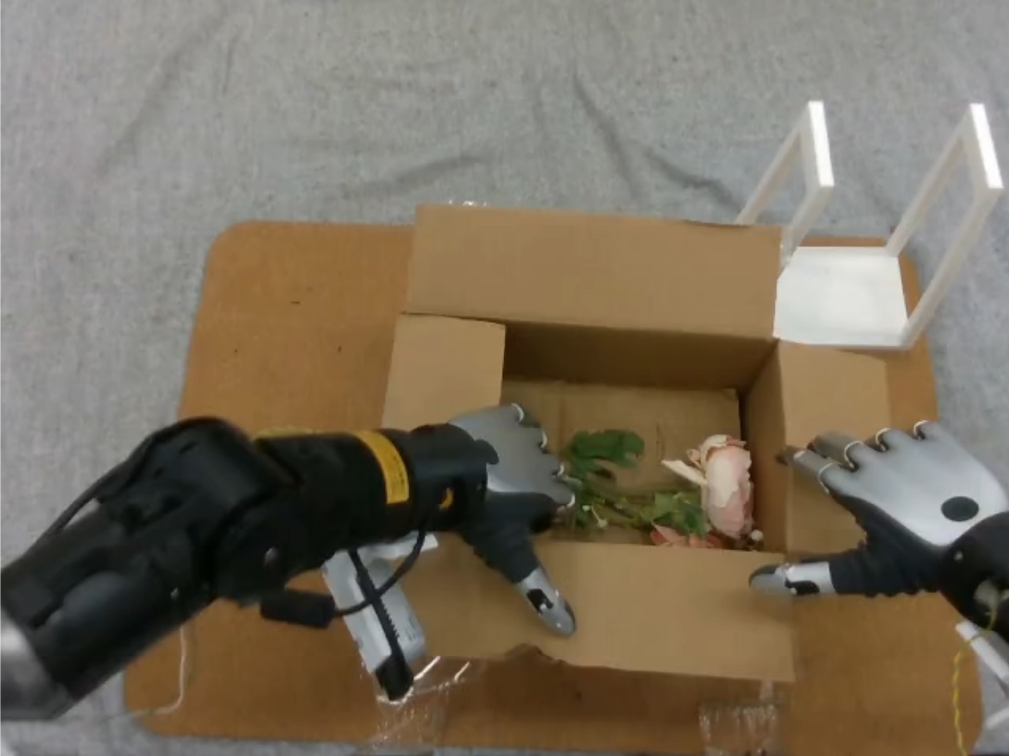} \\
        \scriptsize (a) Raw ego &
        \scriptsize (b) Project+Inpaint &
        \scriptsize (c) \geoformer &
        \scriptsize (d) Robot view
    \end{tabular}
    \caption{\textbf{Qualitative comparison of visual alignment.} (a) the raw ego observation, (b) MoGe reprojection followed by LaMa inpainting, (c) the \geoformer-aligned observation, and (d) the robot-view reference.}
    \label{fig:visual-alignment-qualitative}
\end{figure}

\begin{table}[t!]
    \centering
    \caption{\textbf{Efficiency, image similarity, and downstream performance of visual alignment.}
    Raw uses unaligned ego images; Project+Inpaint applies MoGe reprojection followed by LaMa inpainting; \geoformer is our method.}
    \label{tab:visual-alignment}
    \setlength{\tabcolsep}{3pt}
    \renewcommand{\arraystretch}{1.05}
    \resizebox{\columnwidth}{!}{%
    \begin{tabular}{lccc|cc|cc}
        \toprule
        & \multicolumn{3}{c|}{Alignment Efficiency}
        & \multicolumn{2}{c|}{Image Similarity}
        & \multicolumn{2}{c}{Task I: Open-Box} \\
        \cmidrule(lr){2-4}\cmidrule(lr){5-6}\cmidrule(lr){7-8}
        Method & Mean (ms)$\downarrow$ & P95 (ms)$\downarrow$ & GPU-h / 1M$\downarrow$ & PSNR$\uparrow$ & LPIPS$\downarrow$ & Score$\uparrow$ & Succ.$\uparrow$ \\
        \midrule
        Raw & -- & -- & -- & 11.7187 & 0.7293 & 7.7 & 60.0\% \\
        Project+Inpaint & 162.58 & 194.16 & 45.16 & 12.8213 & 0.7146 & 8.0 & 80.0\% \\
        \specialrule{\lightrulewidth}{\aboverulesep}{0pt}
        \rowcolor{baselinecolor}
        \textbf{\geoformer} & \textbf{7.43} & \textbf{12.97} & \textbf{2.06} & \textbf{13.8749} & \textbf{0.6327} & \textbf{9.5} & \textbf{90.0\%} \\
        \specialrule{\heavyrulewidth}{0pt}{0pt}
    \end{tabular}
    }
\end{table}

\begin{table}[t!]
\centering
\caption{\textbf{Generalization to different package contents.}
All variants originate from the bouquet package-opening policy. Unseen objects are evaluated without object-specific adaptation.}
\label{tab:content-generalization}
\setlength{\tabcolsep}{5pt}
\small
\resizebox{\columnwidth}{!}{%
\begin{tabular}{lccccccccc}
\toprule
& Base & \multicolumn{4}{c}{Seen} & \multicolumn{4}{c}{Unseen} \\
\cmidrule(lr){2-2}\cmidrule(lr){3-6}\cmidrule(lr){7-10}
& Bouquet & Bread & Pouch & Plush & Avg.
& Cable & Tissues & Coke & Avg. \\
\midrule
Score & \textbf{9.5} & 7.0 & 8.8 & 8.6 & \textbf{8.1} & 6.9 & 7.4 & 6.8 & \textbf{7.0} \\
Succ. & \textbf{90.0\%} & 60.0\% & 80.0\% & 70.0\% & \textbf{70.0\%} & 30.0\% & 30.0\% & 40.0\% & \textbf{33.3\%} \\
\bottomrule
\end{tabular}%
}
\end{table}

\paragraph{Processing efficiency}
Table~\ref{tab:visual-alignment} reports end-to-end alignment latency. The Project+Inpaint baseline applies MoGe reprojection followed by LaMa inpainting on every frame. For both methods, we exclude image decoding and model loading from timing and synchronize CUDA after each frame. Mean latency measures average processing time, P95 latency captures tail latency, and GPU-h/1M converts mean latency into the cost of processing one million frames. Raw ego has no alignment module and is therefore excluded from this comparison. \geoformer reduces mean latency from $162.58$~ms to $7.43$~ms and P95 latency from $194.16$~ms to $12.97$~ms. It also reduces the cost from $45.16$ to $2.06$ GPU-hours per million frames, saving $43.10$ GPU-hours ($95.4\%$).

\paragraph{Image similarity}
For the image-similarity results in Table~\ref{tab:visual-alignment}, we compare action-matched ego and robot frames, summarize each episode by its median score, and then average equally across episodes.
PSNR measures pixel-level fidelity; higher is better. LPIPS~\cite{zhang2018perceptual} measures perceptual feature discrepancy; lower is better. Compared with Project+Inpaint, \geoformer improves PSNR by $1.0536$~dB and reduces LPIPS by $11.46\%$. The qualitative comparison in Fig.~\ref{fig:visual-alignment-qualitative} shows the same trend: Project+Inpaint introduces reconstruction artifacts, whereas \geoformer better preserves task-relevant objects and interactions. This stronger alignment also improves the VLA score from $8.0$ to $9.5$ and the success rate from $80.0\%$ to $90.0\%$, outperforming Project+Inpaint.

\subsection{Effect of Human Data Volume}
\label{sec:data-volume}

To investigate how human data volume affects downstream task performance, we vary the amounts of task-specific ego data and glove data. Fig.~\ref{fig:pretraining-and-scaling}(b) shows that both data sources provide their largest measured gains at relatively small data volumes. Performance improves sharply when the first subsets are introduced and then exhibits diminishing returns as additional demonstrations are added. The ego-data curve shows a larger initial change, whereas the glove-data curve saturates earlier. We speculate that task-specific ego data primarily expand task and interaction coverage, while a smaller amount of \glovedata is sufficient to anchor the deployment viewpoint and workspace geometry. The shared saturation suggests that further gains rely more on DAgger-style correction of policy-induced out-of-distribution states than additional human data.

\subsection{Generalization and Transfer}
\label{sec:content-generalization}

\textbf{Objects.} To inspect the object generalization of \sysname, we vary package contents in Task I, adapting the bouquet-based policy to three seen objects with fewer than $100$ additional demonstrations per object.
In Table~\ref{tab:content-generalization}, \sysname achieves an average score of $8.1$ and a $70.0\%$ success rate on the three seen objects. On three unseen objects, the policy also achieves a success rate of $30.0\%$--$40.0\%$ ($33.3\%$ on average). We attribute this promising performance to the broad in-the-wild ego-human prior obtained in Human-to-Robot learning.

\textbf{Embodiments.} To evaluate cross-embodiment transfer, we fine-tune on Tianji Marvin Pro arms with two more arm DoFs in total than \piper, different low-level control, and the same \revo hands. This variant achieves a mean score of $9.2$ and a $60\%$ success rate in Task I, suggesting that our method can transfer well across embodiments.

\section{Conclusion}
\label{sec:conclusion}

We present \sysname for transferring in-the-wild human experience to dual-arm dexterous robots through \geoformer view alignment and progressive human--robot training in a shared robot-native action space. With $538.9$ hours of human experience and less than one hour of robot data per task, our method achieves $90.0\%$--$100.0\%$ success across three long-horizon tasks and $30.0\%$--$40.0\%$ zero-shot success on unseen objects. We hope this work will advance research on scalable VLA models that learn from in-the-wild human experience.

\section{Limitations and Future Work}
\label{sec:limitation}
Learning solely from human demonstrations remains challenging for long-horizon sequential tasks and multi-finger coordination, so the framework still requires real-robot fine-tuning. In addition, \geoformer cannot reconstruct the missing content when head rotations move objects out of the camera's field of view. Finally, we have not yet identified a commercially available higher-DoF hand (\eg, 22 DoF) that jointly meets our requirements for operational stability, durability, thermal management, and affordability, and therefore leave evaluation on such hands to future work.

\section*{Acknowledgments}
We thank the data collection team, especially Jian Yang and Yi Zhang. We are grateful to Peihuan Yang and Zhenkun Lin for their assistance with mechanical design and 3D-printed fabrication. We also thank Huige Tong, Haocheng Zhang, and Yuxing Zhang for their project management support. We further thank the maintenance and after-sales service teams at BrainCo, AgileX Robotics, VIRDYN, and Tianji.

\bibliographystyle{IEEEtran}
\bibliography{main}

\appendices
\linespread{1}\selectfont
\flushbottom
\section*{Appendix}
\setcounter{subsection}{0}
\renewcommand{\thesubsection}{\Alph{subsection}}
\renewcommand{\thesubsubsection}{\arabic{subsubsection}}

The appendix is organized as follows.

\begin{itemize}[leftmargin=*,nosep]
    \item In \hyperref[app:implementation-details]{\textbf{Appendix Section~\ref*{app:implementation-details}}}, we provide details of the hardware and data-collection setups, action alignment, \geoformer training, policy training, and deployment.
    \item In \hyperref[app:experimental-details]{\textbf{Appendix Section~\ref*{app:experimental-details}}}, we describe the experimental protocols, baseline implementations, data composition, additional co-training ablations, action-distribution metrics, and ego--robot pairing procedure.
    \item In \hyperref[app:egowild-details]{\textbf{Appendix Section~\ref*{app:egowild-details}}}, we report data-quality evaluations and detail the annotations, dataset curation, and semantic diversity of EgoWild.
    \item In \hyperref[app:contributions]{\textbf{Appendix Section~\ref*{app:contributions}}}, we list the contributions of each author.
\end{itemize}

\subsection{Implementation Details}
\label{app:implementation-details}

\subsubsection{Hardware and Data Collection Setups}
\label{sec:system-setup}

\begin{figure}[t]
    \centering
    \IfFileExists{Figure/embodiment_teleoperation.pdf}{%
        \includegraphics[width=\columnwidth]{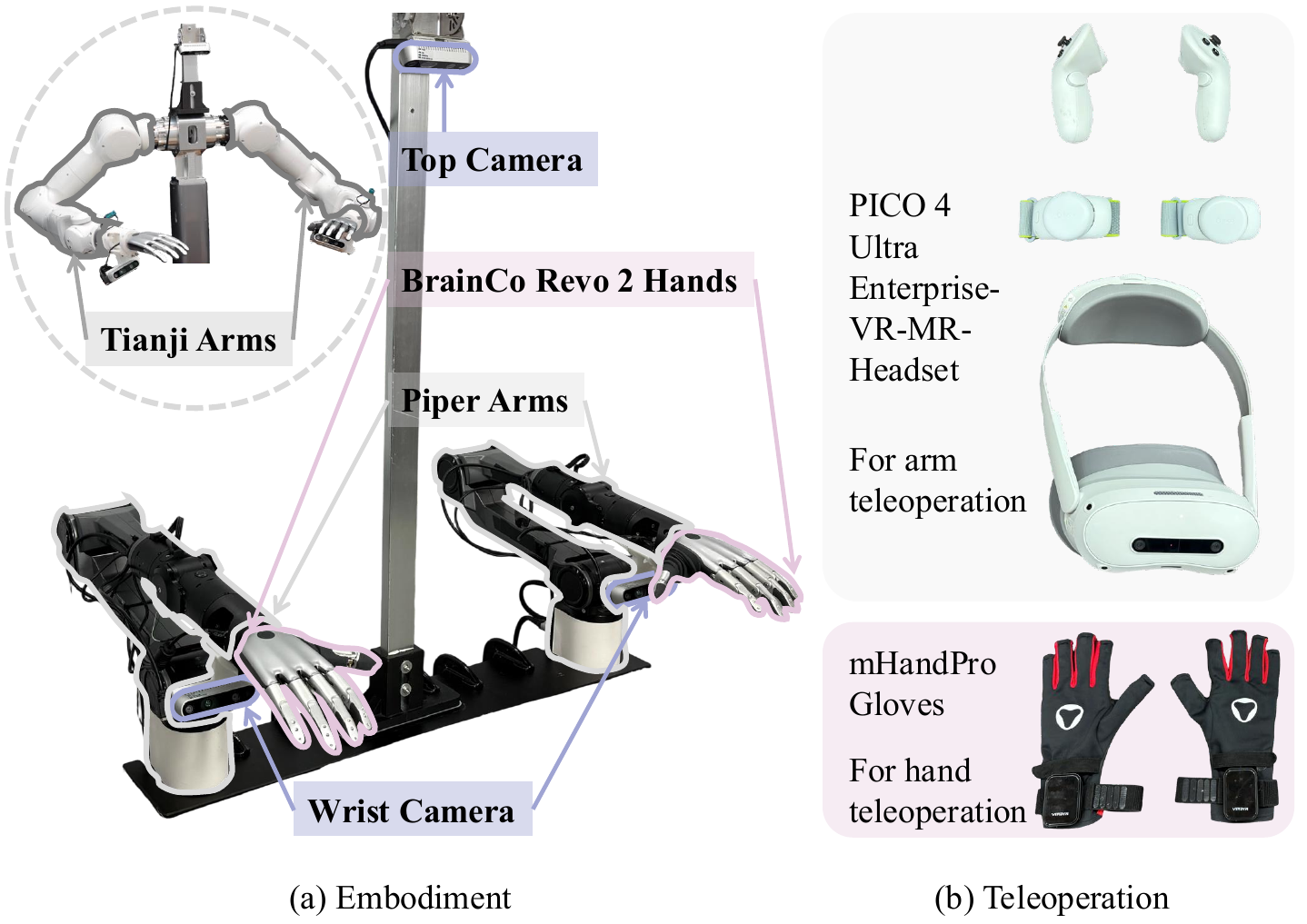}%
    }{%
        \fbox{\parbox[c][3cm][c]{0.9\columnwidth}{\centering
            Embodiment and teleoperation setup\\
            Image pending: \texttt{Figure/embodiment_teleoperation.pdf}}}%
    }
    \caption{\textbf{Embodiment and teleoperation setup.}
    (a) The bimanual robot platform with \piper arms; the upper-left inset uses Tianji Marvin Pro arms with the same \revo hands and camera layout.
    (b) The PICO~4 Ultra Enterprise headset and controllers for arm control, and mHandPro gloves for hand control.}
    \label{fig:embodiment-teleoperation}
\end{figure}

\paragraph{Robot platform}
As shown in Fig.~\ref{fig:embodiment-teleoperation}(a), we use two AgileX \piper arms for our bimanual robot platform. Each $6$-DoF arm carries a tendon-driven $6$-DoF BrainCo \revo hand and a wrist-mounted camera. We mount a top-down camera between the arms to provide a global view. All three cameras are Intel RealSense D435 cameras. The embodiment-transfer variant shown in the upper-left inset uses Tianji Marvin Pro arms while retaining the same hands and camera layout. We control the robot at $30$~Hz.

\paragraph{Ego-human data collection}
For broad and task-specific ego-human demonstrations, collectors interact with objects using their bare hands. PICO's built-in hand tracking records wrist poses and finger motion, while an auxiliary hand tracker improves tracking accuracy during collection. This lightweight setup allows collectors to record their daily activities naturally.

\paragraph{Glove and real-robot data collection}
Fig.~\ref{fig:embodiment-teleoperation}(b) shows the hardware used to collect \glovedata and real-robot demonstrations. A PICO~4 Ultra Enterprise headset and its handheld controllers record arm end-effector poses, while mHandPro gloves measure finger articulation. These measurements control the robot arms and dexterous hands during teleoperation. Before policy training, we process all modalities with the same calibration, IK, hand-retargeting, and command-safety pipeline described in the following subsection.

\subsubsection{Unified Action Alignment}
\label{app:action-retargeting}

The wrist feature $\mathbf{b}^{\mathrm{wrist}}_{j,t}$ specifies the
measured 6-DoF operator pose, while
$\mathbf{b}^{\mathrm{finger}}_{j,t}$ contains the tracked or glove-based
finger channels. Let
$X^{\mathrm{src}}_{j,t}=\operatorname{Pose}
(\mathbf{b}^{\mathrm{wrist}}_{j,t})\in\mathrm{SE}(3)$. Relative motion from the calibrated
configuration at $t_0$ defines the target robot end-effector pose,
\begin{equation}
X^{\star}_{j,t}
=\operatorname{RelPose}\!\left(
X^{\mathrm{src}}_{j,t};
X^{\mathrm{src}}_{j,t_0},
X^{\mathrm{ee}}_{j,t_0}
\right),
\label{eq:relative-retargeting}
\end{equation}
where $X^{\mathrm{src}}_{j,t_0}$ and $X^{\mathrm{ee}}_{j,t_0}$ are the operator
and robot end-effector poses at calibration, respectively, and
$\operatorname{RelPose}$ maps subsequent operator motion to the target
robot pose $X^{\star}_{j,t}$ for side $j$.
The arm and hand components are then generated in parallel,
\begin{equation}
\begin{aligned}
\mathbf{a}^{\mathrm{arm}}_{j,t}
&=\mathcal{H}_{\mathrm{arm}}\!\left(
\mathbf{b}^{\mathrm{wrist}}_{j,t}\right)
:=\operatorname{IK}\!\left(X^{\star}_{j,t}\right),\\
\mathbf{a}^{\mathrm{hand}}_{j,t}
&=\mathcal{H}_{\mathrm{hand}}\!\left(
\mathbf{b}^{\mathrm{finger}}_{j,t}\right),
\end{aligned}
\label{eq:retargeting-details}
\end{equation}
where $\mathcal{H}_{\mathrm{arm}}$ combines calibrated relative-pose
construction with constrained inverse kinematics, and
$\mathcal{H}_{\mathrm{hand}}$ maps calibrated finger features to normalized
hand motor commands. The same calibration conventions, joint constraints,
and channel ordering are used for every demonstration source.

\subsubsection{\geoformer Training}
\label{app:geoformer}

\paragraph{Alignment process}
\geoformer learns from unpaired ego- and robot-view images
$I_{\mathrm{ego}},I_{\mathrm{rob}}\in\mathbb{R}^{3\times H_{\mathrm{img}}\times W_{\mathrm{img}}}$,
where $H_{\mathrm{img}}\times W_{\mathrm{img}}$ denotes the input resolution.
During both \geoformer training and subsequent ego-frame conversion for
policy co-training, each ego image is paired with a robot-view image
randomly sampled from the corresponding task's robot-image pool. The model represents their
geometric difference by an eight-dimensional projective vector
$\mathbf p\in\mathbb{R}^{8}$ and refines it through $N$ predictors with
independent parameters. A parameter vector
$\mathbf p=[p_1,\ldots,p_8]^\top$ is first mapped to the traceless generator
\begin{equation}
    P_{\mathbf p}
    =
    \begin{bmatrix}
        p_3 & p_2 & p_1 \\
        p_6 & -p_3-p_7 & p_5 \\
        p_4 & p_8 & p_7
    \end{bmatrix},
\end{equation}
where $P_{\mathbf p}\in\R^{3\times3}$ has zero trace and encodes the eight
projective parameters. We approximate its matrix exponential using a finite
series with $K_{\mathrm{terms}}$ retained terms:
\begin{equation}
    \mathcal{T}_{\mathbf p}
    =
    \exp(P_{\mathbf p})
    \approx
    \sum_{k=0}^{K_{\mathrm{terms}}-1}
    \frac{P_{\mathbf p}^k}{k!},
\end{equation}
where $K_{\mathrm{terms}}$ is the number of retained series terms and
$\mathcal{T}_{\mathbf p}$ is the resulting homography.
The identity term makes small values of $\mathbf p$ produce transformations close
to the identity, which suits progressive residual refinement.

During training, the initial estimate is
$\mathbf p_0=\sigma_{\mathrm{pert}}\boldsymbol{\epsilon}$, where
$\boldsymbol{\epsilon}\sim\mathcal{N}(\mathbf{0}_8,\operatorname{Id}_8)$. At the $n$-th refinement step, we warp
the robot view with the current estimate,
\begin{equation}
    I_{\mathrm{rob},n-1}^{\mathrm{warp}}
    =\operatorname{Warp}(I_{\mathrm{rob}};\mathcal{T}_{\mathbf p_{n-1}}),
\end{equation}
where $\operatorname{Warp}$ denotes differentiable image warping. The next residual is
predicted from the ego image and the currently warped robot image,
\begin{equation}
    \Delta \mathbf p_n
    =\mathcal G_n(I_{\mathrm{ego}},I_{\mathrm{rob},n-1}^{\mathrm{warp}}),
\end{equation}
where $\mathcal G_n$ is the $n$-th predictor. The estimate is refined as
$\mathbf p_n=\mathbf p_{n-1}+\Delta\mathbf p_n$, and the final robot-to-ego
transformation is $\mathcal{T}=\exp(P_{\mathbf p_N})$.

After a fixed reference mapping from normalized to image coordinates, each
transformed homogeneous point
$\widetilde{\mathbf{h}}=[\widetilde{h}_u,\widetilde{h}_v,
\widetilde{h}_w]^\top$ is normalized by
$u'=\widetilde{h}_u/\widetilde{h}_w$ and
$v'=\widetilde{h}_v/\widetilde{h}_w$. We then use differentiable bilinear
sampling and mark coordinates outside the image support as invalid.

For unpaired training, the final transformation is applied to both the
robot-view image and an all-ones support mask:
\begin{equation}
    I_{\mathrm{rob}}^{\mathrm{warp}}
    =\operatorname{Warp}\!\left(I_{\mathrm{rob}};\mathcal{T}\right),
    \qquad
    M_{\mathcal{T}}
    =\operatorname{clip}\!\left(
        \operatorname{Warp}\!\left(\mathbf{1};\mathcal{T}\right),0,1
    \right),
\end{equation}
where $I_{\mathrm{rob}}^{\mathrm{warp}}$ is the warped robot image and
$M_{\mathcal{T}}\in[0,1]^{H_{\mathrm{img}}\times W_{\mathrm{img}}}$ identifies its valid region. We form
$I_{\mathrm{rob+ego}}=M_{\mathcal{T}}\odot I_{\mathrm{rob}}^{\mathrm{warp}}
+(1-M_{\mathcal{T}})\odot I_{\mathrm{ego}}$, where $\odot$ denotes
element-wise multiplication. The critic evaluates whether the inserted
robot content is geometrically compatible with the ego background,
requiring neither paired images nor explicit 3D reconstruction.

After view-alignment training at the start of Stage~2, we discard the critic,
freeze the predictor, and remove the random perturbation before training
the VLM and action expert. Since training estimates the
robot-to-ego transformation $\mathcal{T}$, its inverse produces the desired
ego-to-robot-aligned observation:
\begin{equation}
    I_{\mathrm{ego}}^{\text{align}}
    =\operatorname{Warp}\!\left(
        M_{\mathcal{T}}\odot I_{\mathrm{ego}};
        \mathcal{T}^{-1}
    \right),
\end{equation}
where $I_{\mathrm{ego}}^{\text{align}}$ is the ego observation aligned toward
the robot view using the inverse homography $\mathcal{T}^{-1}$.
This aligned image is passed directly to the robot policy without scene
reconstruction or content synthesis.

\paragraph{Loss design}
The geometric predictor is trained with the combined objective
\begin{equation}
    \mathcal{L}_{G}
    =\mathcal{L}_{\mathrm{adv}}
    +\lambda_{\mathrm{disp}}\mathcal{L}_{\mathrm{disp}}
    +\lambda_{\mathrm{cov}}\mathcal{L}_{\mathrm{cov}},
\end{equation}
where $\mathcal{L}_{\mathrm{adv}}$ aligns the transformed robot content
with ego-view geometry, $\mathcal{L}_{\mathrm{disp}}$ limits the size of
each refinement update, and $\mathcal{L}_{\mathrm{cov}}$ prevents the valid
warped region from becoming too small or covering nearly the entire image.
The weights $\lambda_{\mathrm{disp}}$ and $\lambda_{\mathrm{cov}}$ balance
the regularizers against the adversarial term.

Let $\mathcal{C}$ denote the critic, which maps an image to a scalar score. The adversarial term
$\mathcal{L}_{\mathrm{adv}}=-\mathbb{E}[\mathcal{C}(I_{\mathrm{rob+ego}})]$
encourages the inserted robot region to be geometrically compatible with the ego-view background.
The displacement term penalizes large residuals across all refinement steps,
$\mathcal{L}_{\mathrm{disp}}=\sum_{n=1}^{N}\mathbb{E}[\|\Delta \mathbf{p}_n\|_2^2]$,
which stabilizes the progressive refinement.
The coverage term controls how much of the warped robot view remains valid. Let $c$ denote the mean of the valid-support mask $M_{\mathcal{T}}$. We use
$\mathcal{L}_{\mathrm{cov}}=[c_{\min}-c]_+ + [c-c_{\max}]_+$,
where $[x]_+=\max(x,0)$. This rules out transformations that retain too little useful content or occupy almost the full frame.
The critic is trained as a Wasserstein critic~\cite{gulrajani2017improved}. For critic updates, we sample a recent composite $I_{\mathrm{rob+ego}}^{\mathrm{hist}}$ from a history buffer and minimize
\begin{equation}
    \mathcal{L}_{\mathcal{C}}
    =
    \mathbb{E}\!\left[\mathcal{C}\!\left(I_{\mathrm{rob+ego}}^{\mathrm{hist}}\right)\right]
    -
    \mathbb{E}\!\left[\mathcal{C}(I_{\mathrm{ego}})\right]
    +
    \lambda_{\mathrm{grad}}\mathcal{L}_{\mathrm{grad}}.
\end{equation}
The first two terms separate generated and real ego images. Specifically, we interpolate
$\widehat{I}=\alpha I_{\mathrm{ego}}+(1-\alpha)I_{\mathrm{rob+ego}}^{\mathrm{hist}}$,
with $\alpha\sim\mathcal{U}(0,1)$, and apply
\begin{equation}
    \mathcal{L}_{\mathrm{grad}}
    =
    \mathbb{E}_{\widehat{I}}
    \left[
        \left(
            \left\|
                \nabla_{\widehat{I}}\mathcal{C}(\widehat{I})
            \right\|_2
            -1
        \right)^2
    \right].
\end{equation}
This promotes unit gradient norm along the interpolation path and stabilizes Wasserstein training.

Finally, a score-margin gate prevents the critic from becoming much stronger
than the predictor. With
$s_{\mathrm{real}}=\mathbb{E}[\mathcal{C}(I_{\mathrm{ego}})]$ and
$s_{\mathrm{fake}}=\mathbb{E}[\mathcal{C}(I_{\mathrm{rob+ego}}^{\mathrm{hist}})]$,
we suspend critic updates when
$s_{\mathrm{real}}-s_{\mathrm{fake}}\geq\tau_{\mathcal{C}}$ while still updating
the geometric predictor.

\paragraph{Optimization}
Each independent predictor uses five stride-$2$ convolutional layers
with widths $(32,64,128,256,512)$. At every scale, the average-pooled
six-channel input is concatenated with the convolutional features. The
resulting representation passes through a $256$-dimensional fully connected
layer and a linear eight-dimensional output head. ReLU is used throughout
except at the output. The fully convolutional critic uses five $4\times4$
stride-$2$ layers with the same channel widths, followed by a linear
$3\times3$ convolution that produces a spatial score map; its hidden layers
use Leaky ReLU with slope $0.2$. Neither network uses normalization. We
initialize weights from zero-mean Gaussian distributions and biases to zero.
The remaining architecture and training settings are summarized in
Table~\ref{tab:geoformer_hyperparameters}.

\begin{table}[t]
    \centering
    \caption{\geoformer architecture and training hyperparameters.}
    \label{tab:geoformer_hyperparameters}
    \small
    \setlength{\tabcolsep}{4pt}
    \begin{tabular}{lc}
        \hline
        \textbf{Hyperparameter} & \textbf{Value} \\
        \hline
        Input resolution & $240\times320$ \\
        Transformation model & Homography \\
        Transformation dimension & $8$ \\
        Refinement steps $N$ & $5$ \\
        Retained exponential-series terms $K_{\mathrm{terms}}$ & $20$ \\
        Training iterations & $5000$ \\
        Batch size & $8$ \\
        Predictor learning rate & $1\times10^{-6}$ \\
        Critic learning rate & $1\times10^{-6}$ \\
        Adam~\cite{kingma2014adam} $(\beta_1,\beta_2)$ & $(0.5,0.999)$ \\
        Predictor initialization std. & $0.02$ \\
        Critic initialization std. & $0.01$ \\
        Initial perturbation $\sigma_{\mathrm{pert}}$ & $0.1$ \\
        $\lambda_{\mathrm{disp}}$ & $0.5$ \\
        $\lambda_{\mathrm{grad}}$ & $10.0$ \\
        $\lambda_{\mathrm{cov}}$ & $1.0$ \\
        Valid-coverage interval & $[0.4,0.8]$ \\
        Predictor updates / iteration & $1$ \\
        Maximum critic updates / iteration & $2$ \\
        Critic margin threshold $\tau_{\mathcal{C}}$ & $0.3$ \\
        Predictor gradient clipping & $5.0$ \\
        \hline
    \end{tabular}
\end{table}

\subsubsection{Policy Training Hyperparameters}
\label{app:policy-training-hyperparameters}

Tables~\ref{tab:stage1-hyperparameters}--\ref{tab:stage3-hyperparameters} summarize the policy training settings for Stage~1 (Human-to-Robot learning), Stage~2 (Human--Robot co-training), and Stage~3 (robot-domain refinement). All stages use NVIDIA A100 GPUs, with $32$ GPUs for Stage~1 and $8$ GPUs each for Stages~2 and~3. All stages use a cosine learning-rate schedule with warmup and an action-chunk horizon of $H=50$; batch sizes are global. Stage~2 (Human--Robot co-training) begins after \geoformer has been trained and frozen. Each Stage~2 batch maintains an approximately $1:1:1$ ratio of task-specific ego, glove, and robot demonstrations.

\begin{table}[t!]
    \centering
    \caption{\textbf{Stage 1 (Human-to-Robot learning) hyperparameters.}}
    \label{tab:stage1-hyperparameters}
    \footnotesize
    \begin{tabular*}{\linewidth}{@{\extracolsep{\fill}} l r}
        \toprule
        \textbf{Hyperparameter} & \textbf{Value} \\
        \midrule
        Learning-rate schedule & Cosine decay \\
        Warmup steps & $5{,}000$ \\
        Peak learning rate & $7.0\times10^{-5}$ \\
        Decay steps & $100{,}000$ \\
        Final learning rate & $7.0\times10^{-6}$ \\
        Training steps & $100{,}000$ \\
        Global batch size & $512$ \\
        Data-loader workers & $8$ \\
        \bottomrule
    \end{tabular*}
\end{table}

\begin{table}[t!]
    \centering
    \caption{\textbf{Stage 2 (Human--Robot co-training) hyperparameters.}
    The data ratio is maintained within each batch.}
    \label{tab:stage2-hyperparameters}
    \footnotesize
    \begin{tabular*}{\linewidth}{@{\extracolsep{\fill}} l r}
        \toprule
        \textbf{Hyperparameter} & \textbf{Value} \\
        \midrule
        Learning-rate schedule & Cosine decay \\
        Warmup steps & $2{,}800$ \\
        Peak learning rate & $3.5\times10^{-5}$ \\
        Decay steps & $50{,}000$ \\
        Final learning rate & $3.5\times10^{-6}$ \\
        Training steps & $50{,}000$ \\
        Global batch size & $128$ \\
        Data-loader workers & $8$ \\
        Batch ratio (ego : glove : robot) & $\approx 1:1:1$ \\
        \bottomrule
    \end{tabular*}
\end{table}

\begin{table}[t!]
    \centering
    \caption{\textbf{Stage 3 (robot-domain refinement) hyperparameters.}}
    \label{tab:stage3-hyperparameters}
    \footnotesize
    \begin{tabular*}{\linewidth}{@{\extracolsep{\fill}} l r}
        \toprule
        \textbf{Hyperparameter} & \textbf{Value} \\
        \midrule
        Learning-rate schedule & Cosine decay \\
        Warmup steps & $2{,}800$ \\
        Peak learning rate & $2.5\times10^{-5}$ \\
        Decay steps & $50{,}000$ \\
        Final learning rate & $2.5\times10^{-6}$ \\
        Training steps & $50{,}000$ \\
        Global batch size & $128$ \\
        Data-loader workers & $8$ \\
        \bottomrule
    \end{tabular*}
\end{table}

\subsubsection{Stable Teleoperation, Deployment, and Online Correction}
\label{app:online-correction}

Reliable robot commands are required during operator-collected demonstrations
and recoveries, as well as when the learned policy controls the robot at
inference time. We therefore stabilize the execution path at three levels:
operator input, policy command execution, and human takeover.

\paragraph{Stable consumer-grade teleoperation}
We use wired communication, packet validation, gap handling, controller-pivot calibration, translation--rotation filtering, rate limits, and temporally initialized IK to suppress tracking jitter and command discontinuities. Tracking loss, excessive IK residuals, or collision detection cause the system to hold the last safe command or stop. The specific calibration and filtering operations are described below.

\paragraph{Mixed Integrated Torque Control}
During contact, a commanded target slightly inside an object can generate large interaction forces, making rigid position tracking unstable. Moreover, when the measured robot state stalls against the environment, supervising the policy with the next proprioceptive state may suppress the operator's continued contact intent. We therefore supervise the policy on commanded actions and execute the resulting joint-position commands using Mixed Integrated Torque Control provided by the Piper arm SDK~\cite{agilex2026pipersdk}:
\begin{equation}
\boldsymbol{\tau}_t=\mathbf{K}_p
\left(\hat{\mathbf q}_t-\mathbf q_t\right)-\mathbf{K}_d\dot{\mathbf q}_t,
\label{eq:mit-control}
\end{equation}
where $\boldsymbol{\tau}_t$ is the joint-torque vector, $\hat{\mathbf q}_t$ is the commanded joint position, $\mathbf q_t$ and $\dot{\mathbf q}_t$ are the measured joint position and velocity, and $\mathbf K_p$ and $\mathbf K_d$ are diagonal proportional and derivative gain matrices. Eq.~\eqref{eq:mit-control} acts as a joint-space spring--damper: the proportional term drives the robot toward the commanded configuration, while the derivative term damps joint motion. This preserves the commanded motion intent while allowing contact-induced spatial errors to be absorbed compliantly.

\paragraph{Streaming action-chunk execution}
Like prior flow-matching VLA deployments~\cite{kai0,rise}, at inference time $t$, the policy predicts an action chunk $\mathbf{A}_t$ of $H$ future actions.
If a new chunk $\mathbf A^\mathrm{new}_{t'}$ is predicted before the active chunk
$\mathbf A^{\mathrm{old}}_{t}$ is exhausted ($t<t'<t+H$), the two chunks
overlap for $N_{\mathrm{ov}} = H+t-t'$ steps. Over this window, we linearly blend the action
$A^{\mathrm{old}}_{t'+r}$ from the active chunk and
$A^{\mathrm{new}}_{t'+r}$ from the new chunk to obtain the executed action $A^\mathrm{exec}_{t'+r}$:
\begin{equation}
A^\mathrm{exec}_{t'+r}
=
(1-\eta_r)A^\mathrm{old}_{t'+r}
+
\eta_r A^\mathrm{new}_{t'+r},
\ \
\eta_r
=
\frac{r+1}{N_{\mathrm{ov}}+1},
\label{eq:action-chunk-blending}
\end{equation}
where $r\in\{0,\ldots,N_{\mathrm{ov}}-1\}$ indexes the overlap window and
$\eta_r$ is the linear blending weight. After the overlap window, the newly
predicted chunk becomes the active chunk. This interpolation avoids abrupt
command replacement between successive policy predictions.

\paragraph{Anchored Delta-Cmd DAgger}
When the policy enters an out-of-distribution state, the operator can intervene to provide a recovery~\cite{kelly2019hgdagger}. An initial pose mismatch between the operator and robot may create a command jump when switching directly to absolute retargeting. We briefly pre-align the operator preview and anchor subsequent motion at the robot command at takeover time $t_{\mathrm{take}}$. For the hand, only the post-takeover change in the operator's finger signal is added to the frozen robot command:
\begin{equation}
\begin{aligned}
\mathbf{m}^{\text{exec}}_{j,t}
&=\mathrm{clip}\!\Bigl(
\mathbf{m}^{\text{rob}}_{j,t_{\mathrm{take}}}
+ \bigl[\mathcal{H}_{\mathrm{hand}}(\mathbf{b}^{\mathrm{finger}}_{j,t})
-\mathcal{H}_{\mathrm{hand}}(\mathbf{b}^{\mathrm{finger}}_{j,t_{\mathrm{take}}})\bigr],\\
&\hspace{5.6em}
\mathbf{m}^{\min},\mathbf{m}^{\max}\Bigr),
\end{aligned}
\end{equation}
where $\mathbf{m}^{\text{exec}}_{j,t}$ is the executed hand command,
$\mathbf{m}^{\text{rob}}_{j,t_{\mathrm{take}}}$ is the robot command frozen
at takeover, and $\mathbf{m}^{\min},\mathbf{m}^{\max}$ are the motor limits.
The difference between the two $\mathcal{H}_{\mathrm{hand}}$ outputs
captures only the operator's post-takeover change.
At $t_{\mathrm{take}}$, the correction is zero, making takeover continuous; afterward, only relative operator motion changes the command. The arm follows the same anchored relative-motion principle, and each recovery is recorded in the shared robot-native action space.

\paragraph{Controller-pivot calibration}
To account for the offset from the controller tracking origin to the wrist
rotation center, we estimate a fixed pivot from $N_{\mathrm{cal}}$
calibration poses:
\begin{equation}
\begin{aligned}
(\boldsymbol{\rho}^{\star},\bar{\mathbf{x}}^{w})
&=\arg\min_{\boldsymbol{\rho},\mathbf{x}}
\sum_{\kappa=1}^{N_{\mathrm{cal}}}\left\|
\mathbf{x}^{c}_{\kappa}+\mathbf{Q}^{c}_{\kappa}\boldsymbol{\rho}-\mathbf{x}
\right\|^{2},\\
\mathbf{x}^{w}_{t}
&=\mathbf{x}^{c}_{t}+\mathbf{Q}^{c}_{t}\boldsymbol{\rho}^{\star},
\end{aligned}
\end{equation}
where $\mathbf{x}^{c}_{\kappa}$ is the measured controller position,
$\mathbf{Q}^{c}_{\kappa}\in\mathrm{SO}(3)$ is its orientation matrix,
$\boldsymbol{\rho}^{\star}$ is the pivot in
controller coordinates, and $\bar{\mathbf{x}}^{w}$ is the estimated
stationary wrist center. This correction prevents pure wrist rotation
from producing false translation.

\paragraph{Pose filtering and command safety}
The corrected translation and measured orientation are filtered and
rate-limited before retargeting and inverse kinematics:
\begin{equation}
\begin{aligned}
\mathbf{x}^{f}_{t}
&=\alpha_x\mathbf{x}^{w}_{t}
+(1-\alpha_x)\mathbf{x}^{f}_{t-1},\\
\mathbf{Q}^{f}_{t}
&=\mathrm{Slerp}(\mathbf{Q}^{f}_{t-1},\mathbf{Q}^{c}_{t},\alpha_Q),\\
\|\Delta\mathbf{x}^{f}_{t}\|
&\leq v_{\max}\Delta t,\qquad
\angle(\Delta\mathbf{Q}^{f}_{t})\leq\omega_{\max}\Delta t,
\end{aligned}
\end{equation}
where $\mathbf{x}^{f}_{t}$ and $\mathbf{Q}^{f}_{t}$ are the filtered position and
orientation, $\Delta\mathbf{x}^{f}_{t}
=\mathbf{x}^{f}_{t}-\mathbf{x}^{f}_{t-1}$ is the translation increment,
and $\Delta\mathbf{Q}^{f}_{t}
=(\mathbf{Q}^{f}_{t-1})^\top\mathbf{Q}^{f}_{t}$ is the relative rotation.
The coefficients $\alpha_x$ and $\alpha_Q$ control smoothing, and
$\mathrm{Slerp}$ denotes spherical linear interpolation. The limits
$v_{\max}$ and $\omega_{\max}$ bound translation and rotation rates over
the control interval $\Delta t$. Dual-arm IK is initialized from the previous command and uses
joint-rate constraints. Stale packets are rejected, short gaps are bridged by holding or interpolating commands, and prolonged tracking loss, excessive IK residuals,
or collision detection freezes the last safe command or stops execution.

\paragraph{Pre-alignment and anchored takeover}
Before takeover, the operator preview is compared with the frozen robot
command at $t_{\mathrm{take}}$. Control switches only after both arm-pose
and hand-command discrepancies fall below fixed safety tolerances.
Corrections follow the anchored Delta-Cmd rule above and are
logged in the robot-native action space.

\subsection{Experimental Details}
\label{app:experimental-details}

\subsubsection{Baseline Models}
\label{app:baseline-details}

\paragraph{\pizero}
This generalist VLA is initialized from an Internet-pretrained $3$B vision--language model and pretrained on Open X-Embodiment together with cross-embodiment dexterous manipulation data collected across eight robot platforms~\cite{pi0}.

\paragraph{\pifive}
This successor to \pizero is designed for open-world generalization through heterogeneous co-training on multi-robot and multi-environment action data, multimodal Web tasks, high-level subtask labels, and verbal-instruction demonstrations~\cite{pi05}.

\paragraph{\textsc{GR00T-N1.7}}
This open $3$B cross-embodiment VLA combines a Cosmos-Reason2-2B backbone with a flow-matching diffusion-transformer action head. It is pretrained on diverse bimanual, semi-humanoid, and humanoid robot demonstrations together with $20{,}000$ hours of EgoScale human video using a shared relative end-effector action representation~\cite{nvidia2026groot17release}.

\subsubsection{Data Composition and Physical Interaction}
\label{app:data-composition}

Table~\ref{tab:task-data-composition} summarizes task-aligned data for each task. Every task uses $1{,}000$ task-specific ego demonstrations, $1{,}000$ \glovedata, and $110$ real-robot trajectories: $100$ expert demonstrations and $10$ DAgger correction trajectories.

\begin{table}[t!]
    \centering
    \caption{\textbf{Task-aligned data composition.}
    Entries report total duration in hours; column headers give demonstration counts.}
    \label{tab:task-data-composition}
    \setlength{\tabcolsep}{3pt}
    \renewcommand{\arraystretch}{1.05}
    \scriptsize
    \begin{tabular}{lccc}
        \toprule
        Task & Ego ($1{,}000$, h) & Glove ($1{,}000$, h) & Robot ($110$, h) \\
        \midrule
        Open-Box & $5.81$ & $5.11$ & $0.64$ \\
        Glue-Figure & $4.16$ & $4.37$ & $0.48$ \\
        Ice-Water & $3.09$ & $3.07$ & $0.51$ \\
        \bottomrule
    \end{tabular}
\end{table}

\subsubsection{Human--Robot Co-Training Ablations}
\label{app:stage2-ablation}

We further ablate the three data sources used during Human--Robot co-training without Human-to-Robot learning. Each variant starts from the same base VLA, follows the same optimization schedule, and retains the same robot-domain refinement; only one Stage~2 source is removed at a time.

\begin{table}[t!]
    \centering
    \caption{\textbf{Data-source ablations during Human--Robot co-training.}
    Downstream results are reported on Task~I: Open-Box.}
    \label{tab:stage2-data-ablation}
    \setlength{\tabcolsep}{4pt}
    \renewcommand{\arraystretch}{1.05}
    \begin{tabular}{ccccc}
        \toprule
        Ego\textsuperscript{TS} & Glove & Robot & Score & Success Rate \\
        \midrule
        \xmark & \cmark & \cmark & 6.6 & 30.0\% \\
        \cmark & \xmark & \cmark & 6.5 & 30.0\% \\
        \cmark & \cmark & \xmark & 5.8 & 20.0\% \\
        \midrule
        \rowcolor{baselinecolor}
        \cmark & \cmark & \cmark & \textbf{7.5} & \textbf{40.0\%} \\
        \bottomrule
    \end{tabular}
\end{table}

\begin{figure}[t!]
    \centering
    \includegraphics[width=\linewidth]{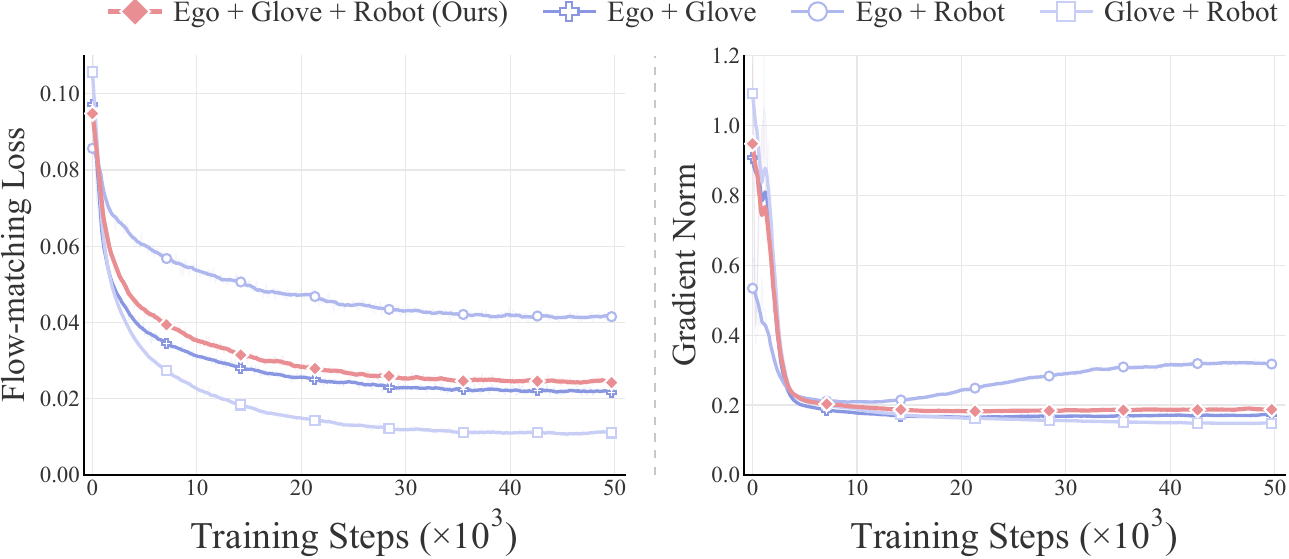}
    \caption{\textbf{Human--Robot co-training optimization with a unified action representation.}
    Each curve corresponds to the full Ego\textsuperscript{TS}+Glove+Robot mixture or one of the three data-source ablations, trained from the same base VLA without Human-to-Robot learning and with the same optimizer, batch size, and schedule. \textbf{Left}: training loss. \textbf{Right}: gradient norm.}
    \label{fig:midtrain_dynamics}
\end{figure}

Removing any Stage~2 source degrades downstream performance, with the removal of robot data producing the largest drop to $5.8/20.0\%$. Removing task-specific ego data or \glovedata yields comparable success of $30.0\%$, but lower scores of $6.6$ and $6.5$, respectively. Fig.~\ref{fig:midtrain_dynamics} shows that directly co-training on task-specific ego and robot data without \glovedata leads to unstable training, whereas adding \glovedata stabilizes optimization. This suggests that glove demonstrations, which combine human manipulation with robot-view observations, help bridge the gap between ego-human and robot data during co-training.

\subsubsection{Action-Distribution Alignment Metrics}
\label{app:action-distribution-metrics}

We use the robot-native action $A_t$ defined in
Eq.~\eqref{eq:action-space-formal} and divide its coordinates into four
groups: left arm, left hand, right arm, and right hand. Before comparing
ego-human and robot actions, we standardize the complete action vector
using statistics pooled from both domains:
\begin{equation}
\widehat A_t
=(A_t-\boldsymbol{\mu})\oslash\boldsymbol{\sigma},
\end{equation}
where $\boldsymbol{\mu}$ and $\boldsymbol{\sigma}$ contain the pooled
mean and standard deviation of each action coordinate, and $\oslash$
denotes element-wise division.

For Fig.~\ref{fig:action-distribution-alignment}(a), we fit PCA jointly
to the standardized ego-human and robot actions and project both domains
onto the same first two principal components. We divide this shared
two-dimensional plane into a $40\times40$ grid. Let
$\mathcal B_{\mathrm{ego}}$ and $\mathcal B_{\mathrm{robot}}$ be the sets
of grid cells occupied by samples from the two domains. Their support
overlap is
\begin{equation}
\mathrm{IoU}_{\mathrm{PCA}}
=\frac{|\mathcal{B}_{\mathrm{ego}}\cap\mathcal{B}_{\mathrm{robot}}|}
{|\mathcal{B}_{\mathrm{ego}}\cup\mathcal{B}_{\mathrm{robot}}|},
\end{equation}
where $|\cdot|$ denotes the number of cells in a set.

For Fig.~\ref{fig:action-distribution-alignment}(b), we take the
Euclidean magnitude of each standardized arm or hand subvector: the
square root of the sum of its squared coordinates. This produces one
nonnegative value measuring how far that group's command is from the
pooled mean. We estimate this magnitude distribution separately for
each group. For any one of the four groups, let
$p_{\mathrm{ego}}(x)$ and $p_{\mathrm{robot}}(x)$ be the normalized
kernel density estimates for ego-human and robot actions at point $x$
on a shared evaluation grid. Their density overlap is
\begin{equation}
\mathrm{OVL}
=\sum_x\min\!\left(
p_{\mathrm{ego}}(x),p_{\mathrm{robot}}(x)\right),
\end{equation}
where the sum runs over the grid and each density sums to one. At every
grid point, the smaller density is the probability mass shared by both
domains; summing this lower envelope gives $\mathrm{OVL}=0$ for
disjoint distributions and $\mathrm{OVL}=1$ for identical
distributions. We apply the same calculation separately to the left
arm, left hand, right arm, and right hand.
All reported alignment values use $10{,}000$ randomly sampled training
frames from each domain.

\subsubsection{Action-Based Ego--Robot Episode and Frame Pairing}
\label{app:visual-frame-pairing}

For the image-similarity evaluation in
Table~\ref{tab:visual-alignment}, we pair ego and robot observations by
how their actions evolve. From each
pair of consecutive actions $A_{t-1}$ and $A_t$, we first compute the
change in every command dimension. We then summarize the changes within
each arm and hand subvector by their root mean square: square the
changes, average them within the subvector, and take the square root.
Each channel is smoothed with a nine-frame moving average and divided by
its temporal $90$th percentile within the episode, so a normalized
value of $1$ represents that channel's $90$th-percentile motion level.
We use a fixed cap of $3$, corresponding to three times the
$90$th-percentile level, so isolated extreme changes do not dominate
the subsequent matching cost. The four normalized channels form the
motion-profile vector $\mathbf v_t\in\R^4$.

\paragraph{Episode-level pairing}
We randomly sample equal numbers of ego and robot episodes. For a
candidate pair, Dynamic Time Warping (DTW)~\cite{sakoe1978dynamic}
aligns their sequences of four-channel motion-profile vectors, using
the Euclidean distance between two vectors as the local cost. DTW can
therefore align similar arm--hand motion patterns performed at different
speeds. The mean of these local distances along the optimal DTW path is
denoted by $\bar c_{\mathrm{DTW}}$. Let $F_{\mathrm{ego}}$ and
$F_{\mathrm{robot}}$ denote the episode lengths in frames. The pair
cost is
\begin{equation}
C_{\mathrm{pair}}
=\bar c_{\mathrm{DTW}}
+\lambda_{\mathrm{dur}}
\left|\log\frac{F_{\mathrm{ego}}}{F_{\mathrm{robot}}}\right|.
\label{eq:app-visual-episode-pairing}
\end{equation}
The first term measures motion-pattern mismatch, while
$\lambda_{\mathrm{dur}}$ controls the penalty for episodes with
different durations. We use
Hungarian assignment~\cite{kuhn1955hungarian} on these costs to obtain
the minimum-cost one-to-one pairing.

\paragraph{Frame-level pairing}
Within each paired episode, we sample every $15$ frames, reducing the
$30$-Hz sequences to approximately $2$~Hz, and run DTW again. Let $i$
and $j$ denote sampled ego and robot frame indices, and let
$u_i^{\mathrm{ego}}$ and $u_j^{\mathrm{robot}}$ denote their normalized
progress from zero at the start of an episode to one at the end. Their
local matching cost is
\begin{equation}
c(i,j)
=\|\mathbf v_i^{\mathrm{ego}}-\mathbf v_j^{\mathrm{robot}}\|_2
+0.2|u_i^{\mathrm{ego}}-u_j^{\mathrm{robot}}|,
\label{eq:app-visual-frame-pairing}
\end{equation}
where $\mathbf v_i^{\mathrm{ego}}$ and
$\mathbf v_j^{\mathrm{robot}}$ are the four-channel motion profiles
defined above, and the second term penalizes differences in episode
progress. We allow a candidate match only when
$|u_i^{\mathrm{ego}}-u_j^{\mathrm{robot}}|\leq0.15$. DTW backtracking
then gives a monotone, possibly many-to-one frame correspondence. Raw
ego, Project+Inpaint, and
\geoformer use exactly the same correspondences and are therefore
evaluated on identical frame pairs.

For each method and metric, we report the unweighted mean of episode-pair median frame scores.

\subsection{Details of EgoWild}
\label{app:egowild-details}

\subsubsection{Data Quality Assessment}
\label{app:data-quality}

\paragraph{Action-annotation accuracy}
We manually audit Qwen3.8-Flash~\cite{qwen2026flash} annotations on $50$ videos ($2{,}926$~s; $906$ human-labeled actions). Of $989$ predicted segments, $97.7\%$ have exact action names and $98.6\%$ have acceptable names (exact or semantically similar).

\paragraph{Temporal-segmentation quality}
We inspect five frames within $\pm 0.6$~s of each candidate cut. A cut is an over-segmentation error only if the hand and object remain in continuous motion, without completion, release, or pause. The confirmed over-segmentation rate is $1.6\%$; the combined segmentation-error rate is $1.7\%$.

\paragraph{Hand-tracking accuracy}
Using the OpenXR middle-fingertip joint on a PICO~4 Ultra, we test unconstrained three-dimensional reaches: a $37$~cm lift followed by a $70$~cm planar motion. Median absolute fingertip-position error is $4.21$~cm without an auxiliary hand tracker versus $0.66$~cm with it. Collectors therefore wear the tracker during recording.

\subsubsection{Dataset Curation and Diversity}
\label{app:broad-ego-corpus}

\paragraph{Episode annotations}
Each episode has one structured JSON annotation with $10$ fields specifying its identifier, start and end times, atomic action, bilingual descriptions, interacting hand, target object class, and noise and cleaning flags for later checks.

\paragraph{Auxiliary metadata and quality filtering}
A recording-level filter uses synchronized body, hand, and camera metadata to find
non-overlapping clips of at least $5$~s with either hand visible
in at least $90\%$ of frames. We remove locomotion intervals lasting at least $1.0$~s with at least
$1.0$~m of horizontal displacement (using head motion as fallback), their $1.0$-s margins, and
quality-control intervals with image blur, malformed hand skeletons, or similar issues. The valid
ratio is the remaining duration divided by the full recording duration. Recordings with a valid
ratio below $0.5$ are labeled low quality and discarded; all others are labeled normal and
retained in full for the next step.

For the second, clip-level filter, a VLM divides each retained recording into timestamped semantic
action segments with descriptions, hand roles, and target objects. We retain segments that are
bimanual and non-noise, exclude Idle, Inspect, and manually rejected intervals, and require each
hand to be active in at least $90\%$ of frames and both hands simultaneously active in at least
$85\%$.

Finally, video, pose, and actions are nearest-neighbor aligned on a shared $30$-Hz timeline. A
frame is invalid if its pose or video timestamp error exceeds $20$~ms or $25$~ms, respectively.
We split each VLM-labeled segment at invalid frames and keep continuous valid parts of at least
$2$~s as final episodes. Each episode keeps the original VLM description as its language
instruction.

\begin{figure}[t!]
    \centering
    \begin{tabular}{@{}c@{\hspace{0.10\columnwidth}}c@{}}
        \includegraphics[width=0.39\columnwidth]{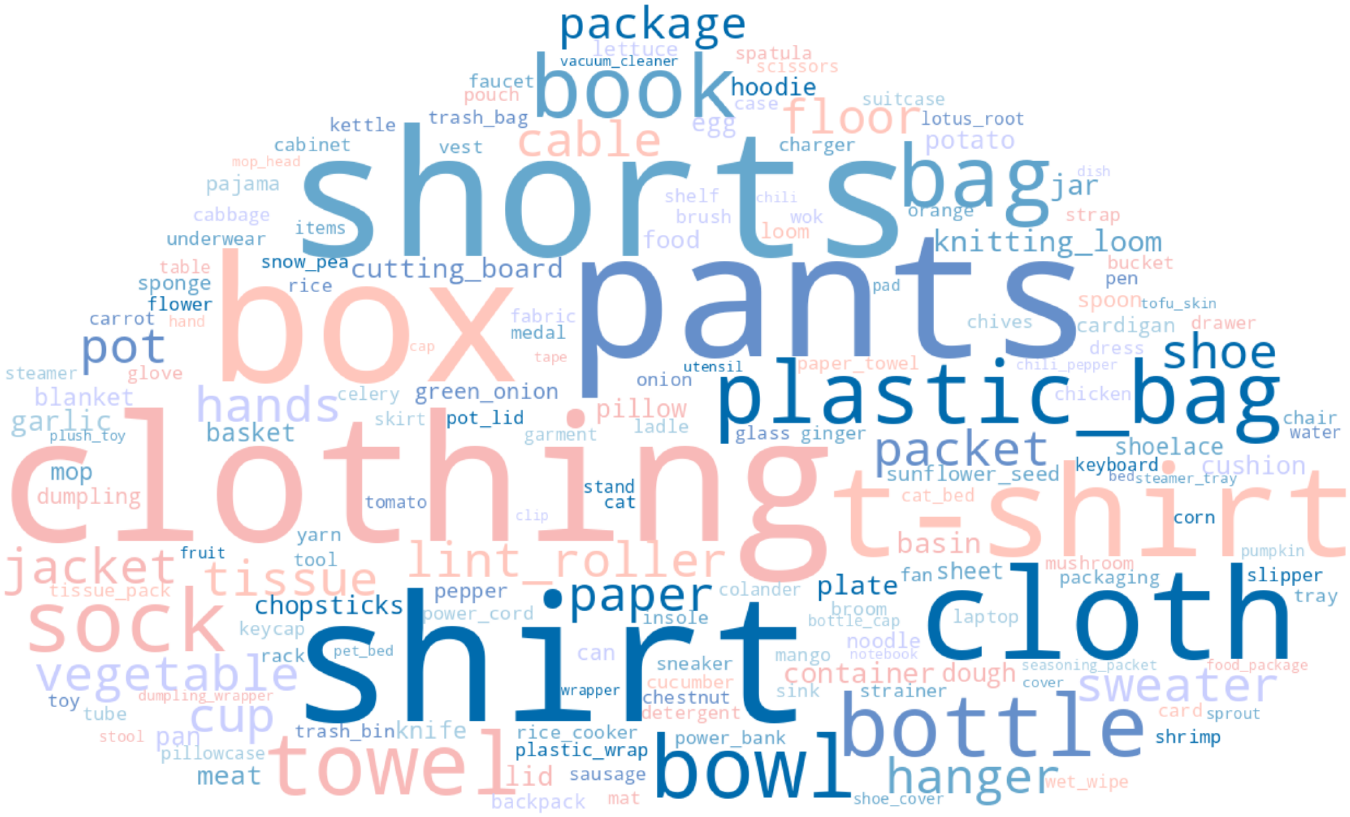} &
        \includegraphics[width=0.39\columnwidth]{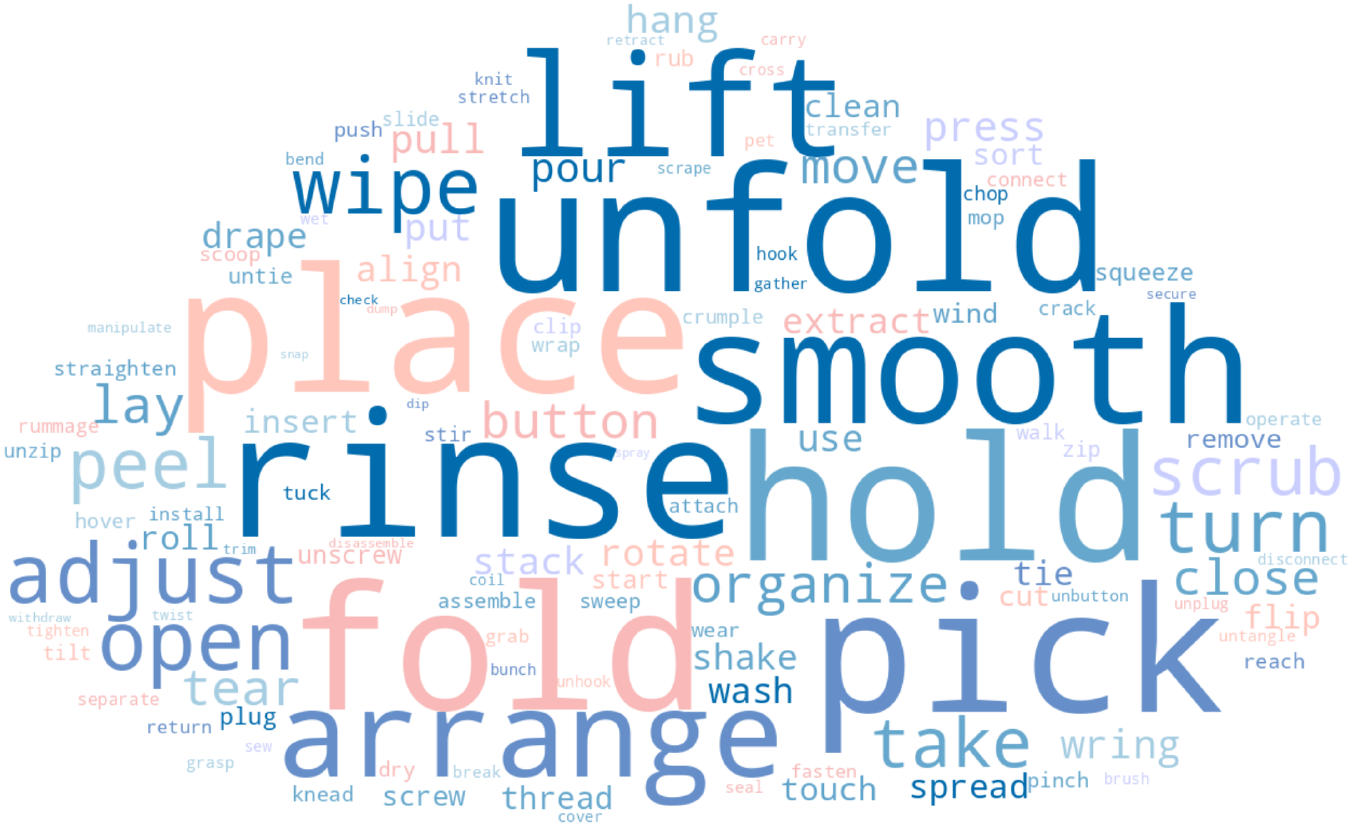} \\
        \scriptsize (a) Objects &
        \scriptsize (b) Skills
    \end{tabular}
    \caption{\textbf{Semantic word clouds for EgoWild.}
    (a) Objects and (b) skills; font size indicates frequency.}
    \label{fig:semantic-wordclouds}
\end{figure}

\paragraph{Qualitative coverage}
To complement the category-level distributions, we visualize semantic coverage using the two word clouds in Fig.~\ref{fig:semantic-wordclouds}. The object cloud in panel (a) spans apparel, household items, and packaging. The skill cloud in panel (b) contains common actions such as \emph{pick}, \emph{fold}, \emph{place}, \emph{hold}, \emph{unfold}, and \emph{rinse}, indicating coverage beyond pick-and-place.

\begin{figure*}[p!]
    \centering
    \includegraphics[height=0.92\textheight,keepaspectratio]{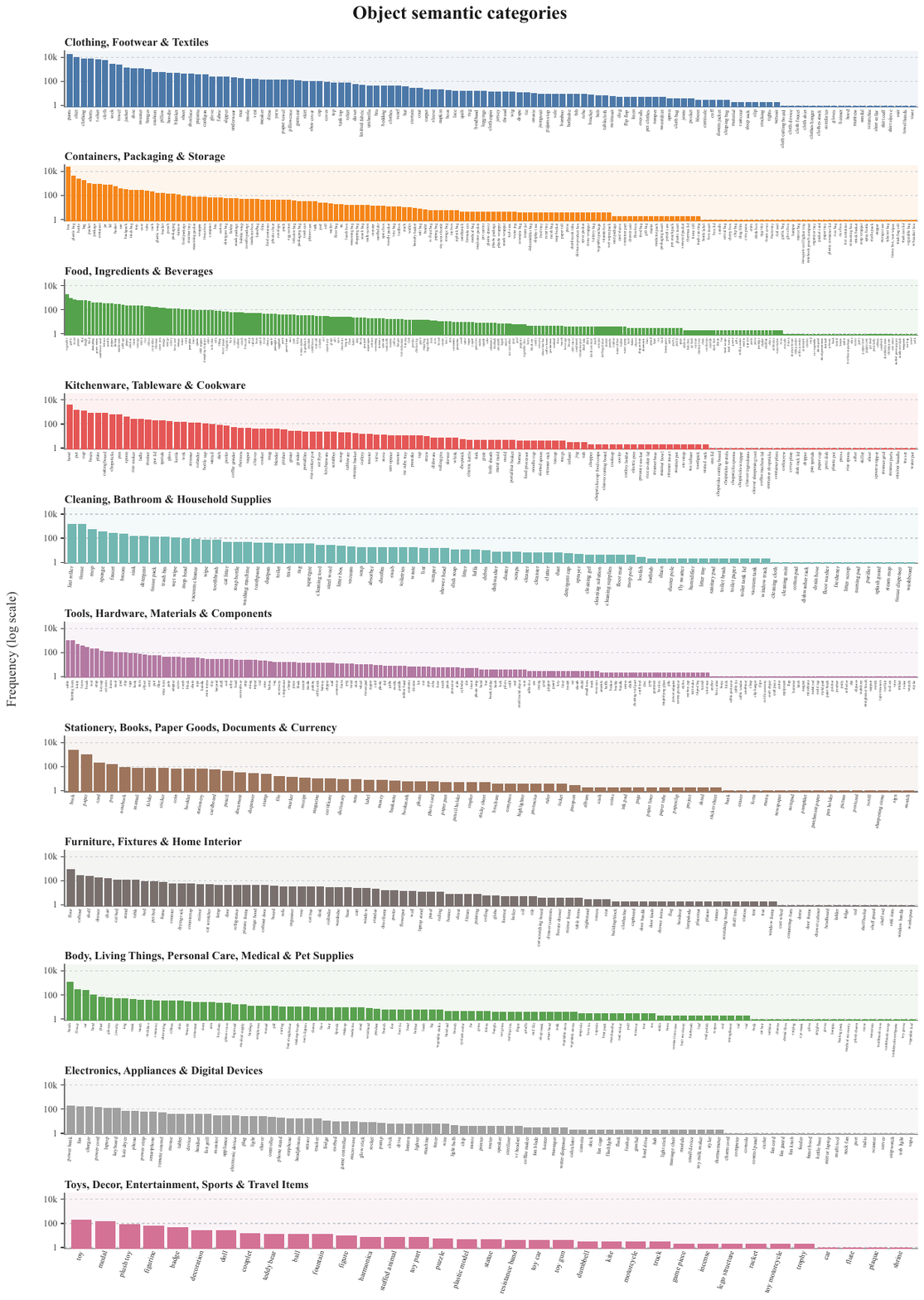}
    \caption{\textbf{Object frequency distributions within broad semantic categories.}
    The $1{,}282$ fine-grained object classes are grouped into broad categories, sorted by frequency within each category, and displayed on a logarithmic scale.}
    \label{fig:object-semantic-categories}
\end{figure*}

\begin{figure*}[p!]
    \centering
    \includegraphics[height=0.92\textheight,keepaspectratio]{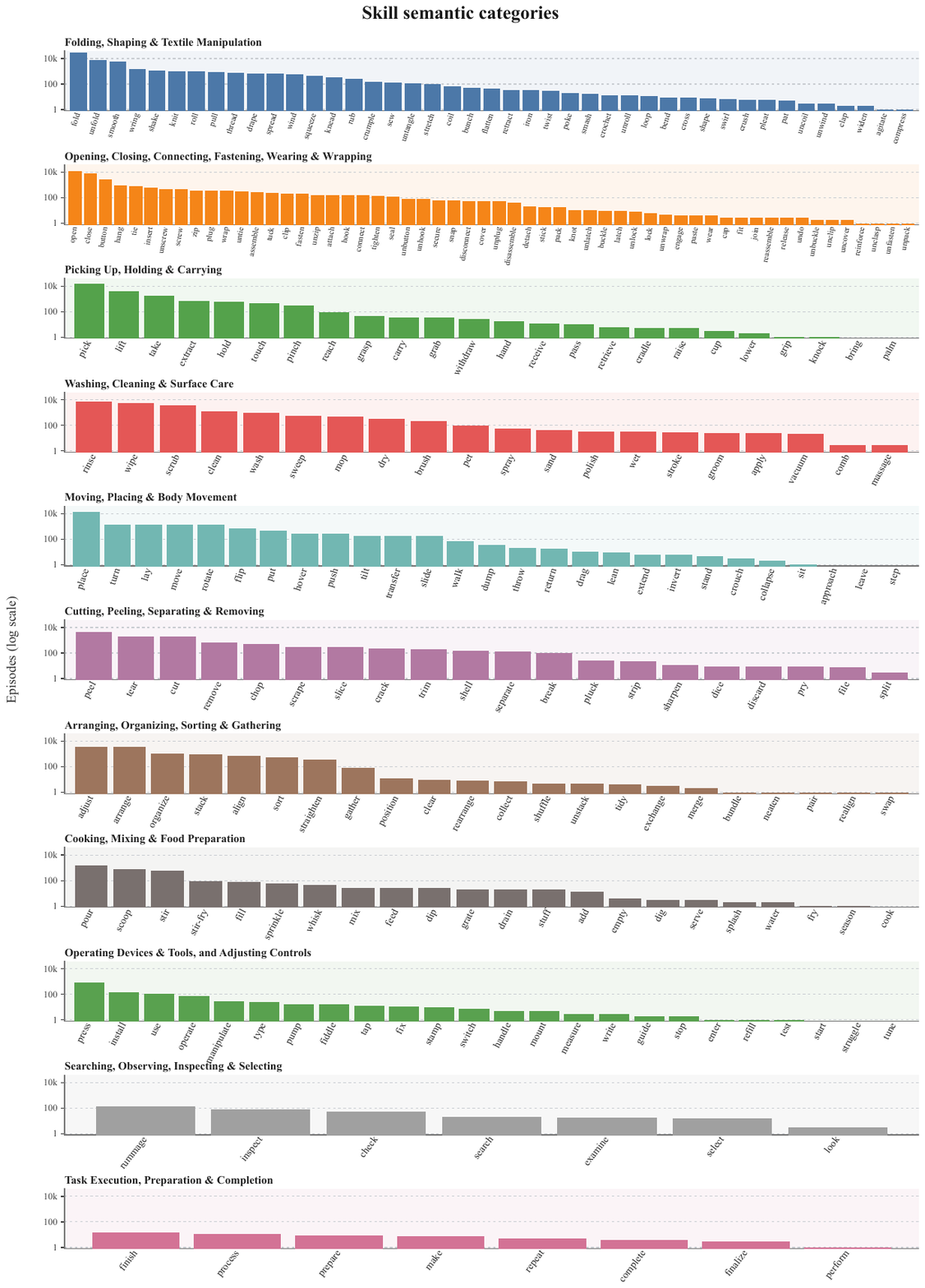}
    \caption{\textbf{Skill frequency distributions within broad semantic categories.}
    The $273$ action-verb lemmas are grouped into broad skill categories, sorted by frequency within each category, and displayed on a logarithmic scale.}
    \label{fig:skill-semantic-categories}
\end{figure*}

\paragraph{Category-level distribution}
Figs.~\ref{fig:object-semantic-categories} and~\ref{fig:skill-semantic-categories} show within-category frequencies for $1{,}282$ object classes and $273$ skill lemmas. Common objects and skills have many annotations, while less frequent entries form a long tail within most categories.

\subsection{Contributions}
\label{app:contributions}
The authors contributed to this work as follows.

\noindent\textbf{Project Lead:} Kunyang Lin.

\noindent\textbf{Core Contributors:} Kunyang Lin, Xutao Wen, Jingxi Lin, Lanyong Lin, and Jiaming Liu.

\noindent\textbf{Learning Algorithm:} Kunyang Lin.

\noindent\textbf{GeoFormer:} Xutao Wen and Kunyang Lin.

\noindent\textbf{EgoWild Dataset:} Jiaming Liu, Kunyang Lin, and Yiduo Li.

\noindent\textbf{Teleoperation and Data Collection System:} Jingxi Lin, Lanyong Lin, and Kunyang Lin.

\noindent\textbf{Training and Deployment:} Kunyang Lin, Xutao Wen, Jingxi Lin, and Lanyong Lin.

\noindent\textbf{Baseline Construction:} Xutao Wen, Lanyong Lin, Kunyang Lin, and Xianchi Chen.

\noindent\textbf{Writing and Illustration:} Kunyang Lin, Xutao Wen, Tianshuo Yang, and Yue Han.

\noindent\textbf{Discussion and Feedback:} Ping Luo, Zhanpeng Zhang, and Tianshuo Yang.

\noindent\textbf{Resources and Academic Supervision:} Ping Luo.

\end{document}

%% file: def.tex
\DeclareMathAlphabet\mathbfcal{OMS}{cmsy}{b}{n}

\def\0{{\bf 0}}
\def\1{{\bf 1}}

\def\eg{\emph{e.g.}}

\def\etc{\emph{etc.}}

\usepackage{amsmath}